\documentclass{article} %
\usepackage{iclr2027_conference,times}

\usepackage{amsmath,amsfonts,bm}

\def\eqref#1{equation~\ref{#1}}

\def\1{\bm{1}}

\DeclareMathAlphabet{\mathsfit}{\encodingdefault}{\sfdefault}{m}{sl}
\SetMathAlphabet{\mathsfit}{bold}{\encodingdefault}{\sfdefault}{bx}{n}

\usepackage{hyperref}
\usepackage{url}
\usepackage{enumitem}
\usepackage{amsmath}
\usepackage{mathtools}
\usepackage[table]{xcolor}
\usepackage{pgf}
\usepackage{collcell}
\usepackage{multirow}
\usepackage{wrapfig}
\usepackage{placeins}
\usepackage[bottom]{footmisc}
\usepackage{xurl}
\usepackage{cleveref}

\DeclarePairedDelimiter{\set}{\{}{\}}

\definecolor{heatLo}{HTML}{c7101b}  %
\definecolor{heatMid}{HTML}{F5F0E1}
\definecolor{heatHi}{HTML}{6BE59A}  %
\definecolor{heatHiTxt}{HTML}{449664}

\newcommand{\heatTint}{60}

\newcommand{\heatbg}[3]{%
  \pgfmathsetmacro{\heatT}{max(0,min(1,((0#3)-(#1))/((#2)-(#1))))}%
  \ifdim\heatT pt<0.5pt
    \pgfmathsetmacro{\heatP}{200*\heatT}\edef\heatCol{heatMid!\heatP!heatLo}%
  \else
    \pgfmathsetmacro{\heatP}{200*\heatT-100}\edef\heatCol{heatHi!\heatP!heatMid}%
  \fi
  \edef\heatDo{\noexpand\cellcolor{\heatCol!\heatTint!white}}\heatDo}
\newcommand{\dg}{\ensuremath{^{\dagger}}}
\def\heatparse#1\dg#2\heatstop{\def\heatnum{#1}}
\newcommand{\heatcell}[3]{%
  \heatparse#3\dg\heatstop
  \heatbg{#1}{#2}{\heatnum}#3}
\newcommand{\heatcellmath}[3]{%
  \heatparse#3\dg\heatstop
  \heatbg{#1}{#2}{\heatnum}$#3$}

\newcolumntype{H}[2]{>{\collectcell{\heatcell{#1}{#2}}}r<{\endcollectcell}}
\newcolumntype{M}[2]{>{\collectcell{\heatcellmath{#1}{#2}}}r<{\endcollectcell}}

\newcolumntype{P}{H{0.0}{1.0}}  %
\newcolumntype{E}{M{-1.0}{1.0}} %
\newcolumntype{F}{M{-0.25}{1.0}}  %
\newcolumntype{G}{@{}p{10pt}@{}}  %

\title{When Generative Models Act as Simulators}

\author{%
\makebox[\dimexpr\textwidth-2\tabcolsep\relax][c]{Sevan Brodjian \quad Michael Hobley \quad Pietro Perona}\\[2pt]
\makebox[\dimexpr\textwidth-2\tabcolsep\relax][c]{\small California Institute of Technology \quad \texttt{\{sbrodjia,mahobley,perona\}@caltech.edu}}%
}

\colorlet{revcolor}{black}
\newcommand{\rev}[1]{\textcolor{revcolor}{#1}}
\newenvironment{revblock}{\color{revcolor}}{}

\newcolumntype{g}{@{}p{5pt}@{}}  %

\iclrfinalcopy %
\begin{document}

\maketitle
\lhead{}
\renewcommand{\headrulewidth}{0pt}
\vspace{-2pt}
\begin{abstract}
A simulator summarizes an environment in a state that users can read and change. Generative models learn representations of the world without explicit specification, but offer no built-in way to edit that world's state. We investigate the conditions under which a generative model \rev{acts} as a simulator through \textit{editability}: whether changing a model's latent representation produces the next-step outputs we get by editing a simulator's state. We test two kinds of \rev{state} mappings, both \rev{fit} from \rev{latent states} paired with simulator states. In one, we train probes to decode the simulator state from the model, then steer along them. In the other, we learn the \rev{simulator-to-latent} mapping directly with an inverse map, which can succeed even \rev{when} steering fails. We study both in two contrasting environments: the discrete board game Othello, and Rayworld, our own continuous world of moving discs seen through partial renderings, closer to physical scenes and video. We find that which mapping succeeds depends strongly on the environment, \rev{in ways decodability does not predict}. In Othello, \rev{each simplification of the rules lowers editability.} In Rayworld, probes fail to steer along the simulator's continuous state, yet an inverse map sets the model from that state anyway. \rev{Editability across layers suggests that the Othello model compresses its history into a board partway through the network, while the Rayworld model shifts its recent frames and then refines a rendering of the next frame. The inverse map succeeds by approximating this partial rendering. We relate both outcomes to the causal state of computational mechanics, the minimal summary of its history a model must keep to predict optimally, which may not align with the simulator. Guided by this view, we change how the Rayworld model is trained and make its state more simulator-like.}%
\footnote{Code at \href{https://github.com/SevanBrodjian/generative-models-as-simulators}{\nolinkurl{github.com/SevanBrodjian/generative-models-as-simulators}}.}
\end{abstract}

\section{Introduction}

\begin{figure}[b]
\centering
\includegraphics[width=0.975\linewidth]{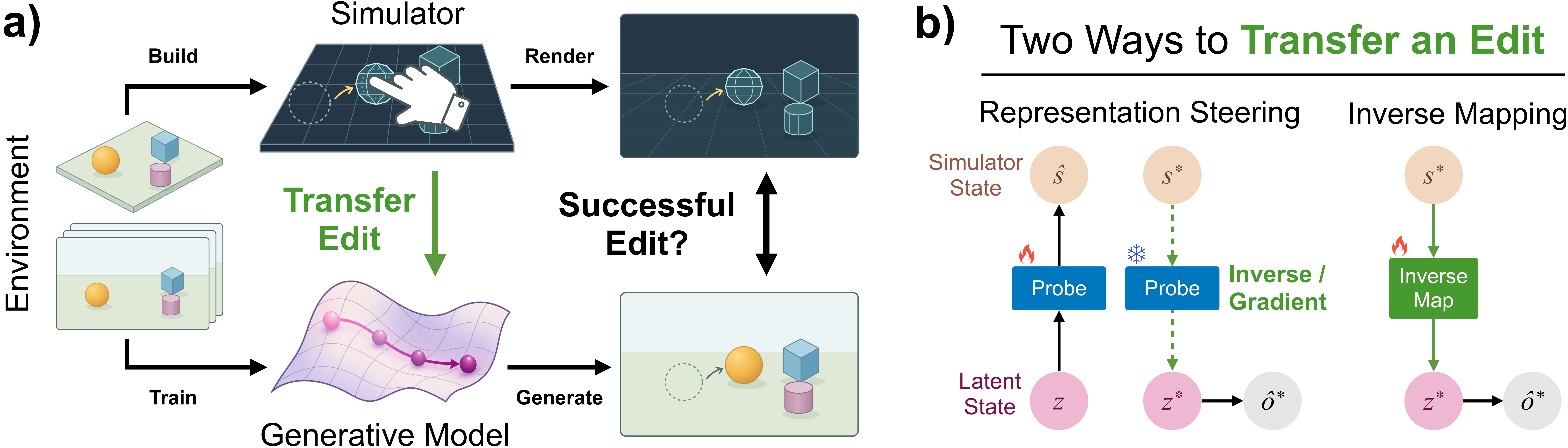}
\caption{We examine when a generative model's latent state can be set from interpretable simulator variables. (a) We train a generator on an environment's observations alone, then test whether an edit to a ground-truth simulator transfers into it, so its generation matches the simulator's rendering. (b) In representation steering, we train a probe to read out simulator state, then invert it or follow its gradients to steer the latent state. In inverse mapping, we train a map from simulator state to latent state. Neither approach is trained on the observations its edits should produce. See \cref{fig:experimental_setup} for details.}
\label{fig:teaser}
\end{figure}

A simulator represents an environment by holding a state, advancing it by a rule, and rendering it into observations. In conventional simulators, this state is explicit and interpretable, allowing it to be set while preserving commitments to that environment and its rules \citep{todorov2012mujoco}. Yet what these systems can express is constrained by what has been specified. Generative models, by contrast, learn from observations and can supply plausible details where that specification is incomplete \citep{brooks2024sora,bruce2024genie}. If an observer has seen only half of a room and then turns around, a generative model can propose what lies behind them, consistent with what they have seen \citep{lu2025genex}. Their internal representations, however, offer no built-in means to set the environment's state; one must be found or engineered \citep{shen2020interfacegan,tewari2020stylerig}.

Combining the two would provide interpretable control with learned detail. In forecasting \citep{lam2023graphcast} and planning \citep{hafner2025dreamerv3}, this could pair predictive accuracy with the ability to specify precise conditions and counterfactuals. In virtual environments, a generative model could continuously extend the scene while a simulator keeps its state and rules persistent. Both require \textit{editability}: the ability to set a model's latent state so that its next prediction follows from a target simulator state. Crucially, editability is a question about the simulator-like functionality we can access, not about whether the model represents or advances state as a conventional simulator does.

We investigate which conditions make a variable of the environment editable within an unsupervised generative model. Prior work shows that decodability alone does not reveal editability \citep{hazineh2023linear}. To avoid confounds from the architecture, training, or evaluation, we hold all three fixed and vary only the training environment. Changing the rules governing a variable, or the observations through which it is revealed, then isolates their effect on its editability in the learned model. 

We study two classes of environment: Othello, whose discrete board state is fully determined by its move history and rules \citep{li2023othello}, and Rayworld, whose continuous disc positions are rendered into partial 1D observations, closer to physical scenes and video. We vary the game rules in Othello and the observation process in Rayworld, and also compare continuous and categorical descriptions of the discs' state. Both environments have known ground truth simulators, so the correct response to each edit can be checked directly. We score each edit by whether it moves the predicted observation toward what the environment produces after the edit, and by its overall fidelity.

Given observations paired with their simulator states, we compare two ways of editing a model. The first, representation steering, comes from mechanistic interpretability. Probes (linear or MLP) are trained to decode the state from the model's latent representations, revealing how the model encodes it, and editors then move the latent state along the probes' directions \citep{li2023othello,nanda2023linear}. A failed steering edit, however, shows only that the probes expose no control, not that the model cannot act as a simulator. The second, inverse mapping, tests this directly. A map trained on the same data, from simulator state to latent state, overwrites the latent state at a single layer. The model's remaining layers must then carry that write through to its output. If it succeeds where steering fails, we can still drive the model with simulator states, even if the model does not represent them as the simulator does.

Every model decodes its state well, yet in Othello \rev{each simplification of the rules lowers editability.} This holds even for a variant whose next move still depends on tracking the full board, where steering along the probes already degrades sharply. In Rayworld, the probes fail to steer unless their target is discretized, and steer best when it mirrors the observation, suggesting the model never forms a single abstract representation of where the discs are. Yet the inverse map succeeds from continuous positions, even though it predicts the model's latent state poorly. \rev{Comparing editability across layers shows how each model works. The Othello model compresses its history into a board at its middle layers. The Rayworld model appears to shift its recent frames and then refine a rendering of the next frame, which the inverse map approximates. Using the causal states of computational mechanics, we relate both outcomes to whether prediction pressures a model to compress its history into the simulator's state, and use this to predict a change that raises Rayworld's editability.}

Our main contributions are as follows:
\begin{enumerate}[
label=(\arabic*),
leftmargin=*,
align=left,
itemsep=0.25em,
topsep=0pt,
parsep=0pt,
partopsep=0pt
]
\item \textbf{An evaluation framework for editability across diverse environments;}
\item \textbf{\rev{Evidence that each simplification of Othello's rules lowers editability;}}
\item \textbf{\rev{An analysis of editability through each network, showing how models of each environment compute their predictions and suggesting why a model might bypass a simulator-like state.}}
\end{enumerate}

\section{Background and Related Work}

\paragraph{World Models and Simulators.}
World models range from learned environments for agent training \citep{ha2018worldmodels} to video generators presented as world simulators \citep{brooks2024sora,bruce2024genie,nvidia2025cosmos}. Whether they represent environment state and use it in prediction \citep{li2025definition} is distinct from whether that state can be set. \citet{vafa2024evaluating} test whether models predict the same futures for the same state, and different futures for different states. For video generators, \citet{kang2025physical} find accurate physical predictions within the training distribution but poor extrapolation, while What-If World tests whether prompted changes to physical variables produce the expected consequences \citep{cai2026whatif}.

Controllable generation often relies on additional structure: trained conditioning interfaces \citep{zhang2023controlnet,wang2024motionctrl,nvidia2025transfer}, geometric manipulation of latent features for camera control without further training \citep{hou2026camera}, or generative priors that infer material properties for an explicit physical simulator \citep{zhang2024physdreamer}. The available edits thus depend on the structure provided. We instead ask whether a model trained only to predict observations can accept simulator state edits through its learned representations, producing the observations expected from a target simulator state.

\paragraph{Probing and Steering Learned Representations.}
In mechanistic interpretability, a common approach to studying learned world models is to decode environment variables and then intervene on their representations to test whether the model uses them. In Othello, \citet{li2023othello} decode board states from transformers trained on move sequences and use probe gradients to guide predictions toward edited boards. Representing token ownership as mine/theirs permits linear decoding and edits through vector arithmetic \citep{nanda2023linear}, and similar interventions produce moves legal under edited chess boards \citep{karvonen2024chess}. Yet accurate board decoding can coexist with ineffective edits in late layers and shallow Othello models \citep{hazineh2023linear}. \rev{\citet{chawla2026metaothello} study how one transformer represents several Othello variants, and find a largely shared board representation, through which probes trained on one variant can intervene on another. We instead ask how a single variant's rules decide whether its board can be edited at all.}

Decoding accuracy must be interpreted with care, since a probe can itself learn the structure being recovered \citep{hewitt2019probes}. Amnesic probing tests whether a model uses certain decoded information by removing it from activations and measuring the effect on behavior \citep{elazar2021amnesic}. In video encoders, \citet{joseph2026physics} edit motion representations and use a separate probe to check whether they now indicate the target direction. \citet{esmati2026invisible} study whether video-model activations distinguish physically plausible from implausible events. In Atari models, \citet{zhang2026probing} also shows that interventions along probe directions change predictions. Because our simulators can be edited directly, we can instead compare a model's edited output with the output of the corresponding simulator edit. As in image editing \citep{shen2020interfacegan} and language-model steering \citep{zou2023repe,rimsky2024caa}, success also depends on preserving properties outside the target \citep{canby2025reliable}. Our metrics measure both the targeted change and prediction quality.

A direct mapping into latent space can also set a model's state, without first finding a steerable representation. StyleRig maps face parameters and the current latent into StyleGAN using editing losses \citep{tewari2020stylerig}, while our map uses state alone. Closer to our setting, \citet{liu2026carriers} compare edited rollouts against simulator counterfactuals, but need counterfactual trajectories to train both model and edits. Our inverse mapping is fitted only on ordinary state--activation pairs, and we test whether it produces the intended output when probe-derived editors cannot.

Causal abstraction tests whether interventions on a model's internal representations correspond to interventions on variables in a higher-level causal model \citep{geiger2021causal}, typically by swapping activations between inputs and, in distributed alignment search, learning the subspace where such swaps succeed \citep{geiger2025foundation,geiger2024das}. We share this intervention-based criterion, but our high-level model is the simulator we want to drive the model with, not a hypothesis about its computation. Our editors are fit only to paired latent and simulator states, without counterfactuals or optimization for intervention success, which lets us test whether decodable representations are also editable.

\section{Experimental Setup}
\label{sec:experimental_setup}

\begin{figure}[t]
\centering
\includegraphics[width=0.9\linewidth]{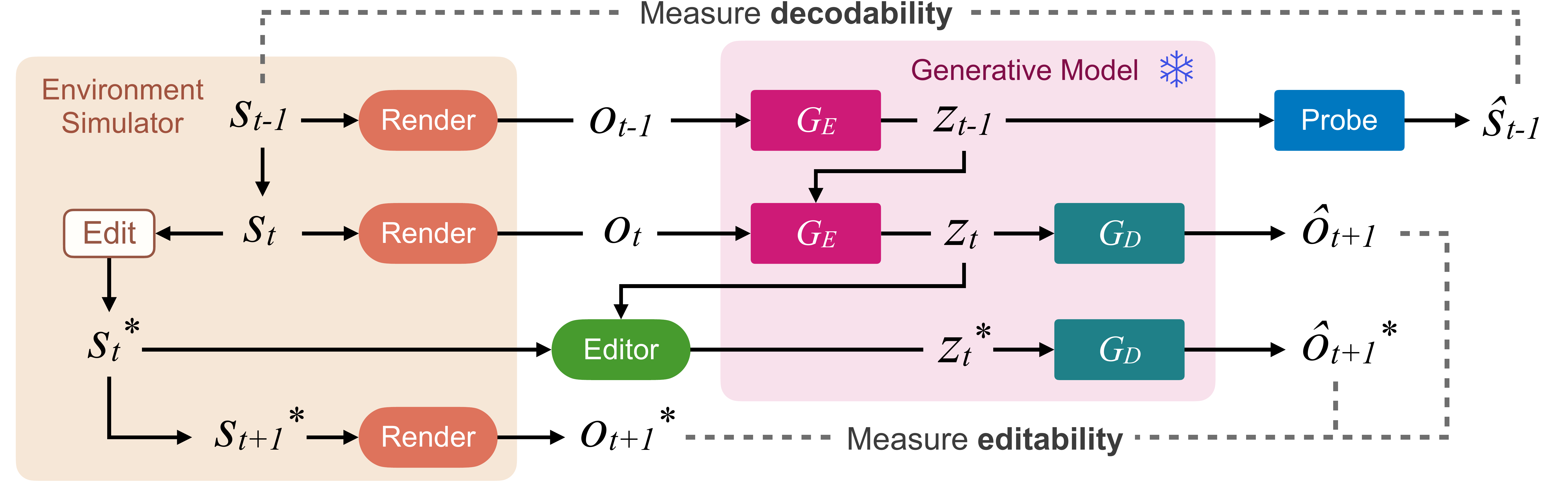}
\caption{Overview of our evaluation procedure. The environment simulator renders state $s_t$ into observation $o_t$. The frozen generative model encodes the observations into latent state $z_t$ with $G_E$, and $G_D$ maps it to the next-step prediction $\hat{o}_{t+1}$, where $z_t$ holds the residual activations at every point and $G_D$ reads the last. We measure \textbf{decodability} by training probes to recover simulator state from the latent state. To measure \textbf{editability}, we edit the simulator state to obtain a target $s_t^*$, use an editor to produce a corresponding latent state $z_t^*$ and pass it through $G_D$ for the edited prediction $\hat{o}_{t+1}^*$. We compare this prediction with $o_{t+1}^*$, rendered from the edited state one step forward, using the unedited observation $o_{t+1}$ (not shown) and prediction as references (\cref{eq:editability_formulae}).}
\label{fig:experimental_setup}
\end{figure}

Our experiments hold the model and its analysis fixed while the environment changes, so a difference between two rows of our tables is a fact about the environment. We use a common scheme for both environments, shown in \cref{fig:experimental_setup}. A simulator holds a state $s_t$ and renders it into an observation $o_t$, a move in Othello and a frame in Rayworld. The encoder $G_E$ maps $o_t$ and earlier latent states $z_{1:t-1}$ to $z_t$, the residual stream at every point, and the decoder $G_D$, the output layer, reads the last point to predict $\hat{o}_{t+1}$. An edit changes $s_t$ to a target $s_t^*$, an editor proposes a matching latent state $z_t^*$, and $G_D$ maps it to $\hat{o}_{t+1}^*$. The reference is $o_{t+1}^*$, rendered from $s_t^*$ advanced one step, with the unedited $o_{t+1}$ as a baseline. We vary the rules in Othello and what the observation reveals in Rayworld.

\subsection{Othello}
\label{sec:othello_experiments}
Othello is played on an $8 \times 8$ board. Players alternate placing tokens, and a legal move must enclose a line of opponent tokens, which then flip to the mover's color. We follow the setup of \citet{li2023othello}, whose generator draws every move uniformly from the legal set, so the optimal prediction is known exactly: a uniform distribution over the legal moves (\cref{app:predict_skill}). Following \citet{nanda2023linear}, instead of using black and white we describe the board as mine and theirs relative to the player about to move, the form in which it is linearly decodable.

The simulator state $s_t$ is the board after each move, with every cell being empty, mine, or theirs. An edit flips the color of one token after a real 20-move opening and is judged against the legal moves of the board before and after the flip, according to each variant's own placement rules. The flipped board need not be reachable by any legal game, a property inherited from the implementation of \citet{li2023othello}. We split our results by reachability in \cref{app:legal_vs_illegal_othello}.

\begin{figure}[t]
\centering
\includegraphics[width=0.8\linewidth]{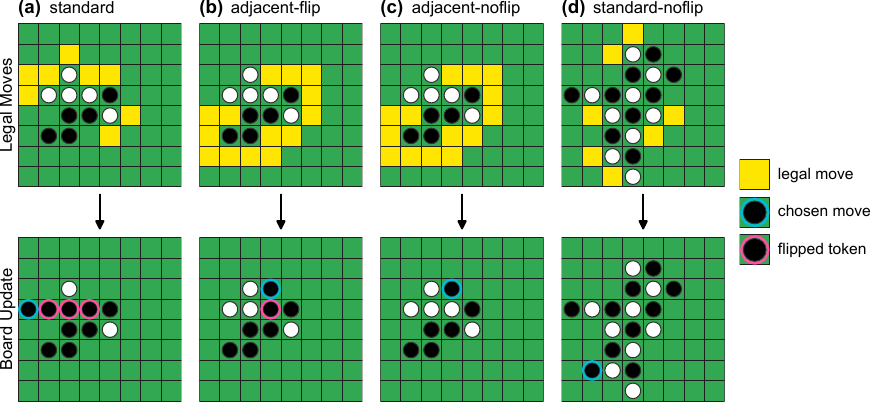}
\caption{(a) In \texttt{standard} Othello, players alternate placing tokens that must enclose opponent tokens, which then flip color. (b) \texttt{adjacent-flip} instead allows placement anywhere adjacent (or diagonal) to one's own color, (c) \texttt{adjacent-noflip} also turns flipping off, and (d) \texttt{standard-noflip} keeps the original placement without flipping, so every game fills the same checkerboard in a different order. In all examples it is black's turn to move. See \cref{sec:othello_experiments}.}
\label{fig:othello_and_variants}
\end{figure}

\paragraph{Othello Variants.}
We introduce three variants of Othello. One modifies where a token may be placed, another removes the flipping of enclosed tokens, and the third does both. \rev{Each change makes the game simpler to play. Adjacent placement needs only a local color check, and without flips a token never changes once placed.} In \texttt{adjacent-flip} a token may be placed anywhere adjacent to a token of its own color, including diagonally, with flipping active. Flips still change the legal moves but become rarer, at 0.27 flipped tokens per move against 2.2 in the \texttt{standard} game. \texttt{adjacent-noflip} uses the same placement rule with flipping off, so a placed token never changes color. \texttt{standard-noflip} keeps the standard placement rule without flipping, so each game fills the same checkerboard in a different order, and a token's color is fully determined by the cell it occupies. \Cref{fig:othello_and_variants} illustrates the setup and its variants.

\subsection{Rayworld}
\label{sec:rayworld_experiments}
\begin{figure}[t]
\centering
\includegraphics[width=0.925\linewidth]{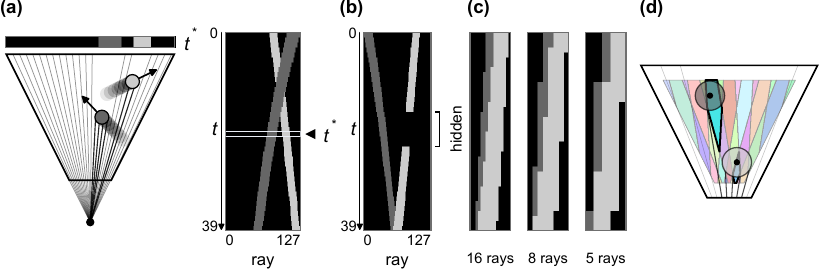}
\caption{(a) In \texttt{standard} Rayworld, two discs move at constant velocity in a 2D frustum without colliding. Each frame is rendered into a 1D observation by casting rays into the frustum, shown stacked over time. (b) In \texttt{blink}, discs vanish for signaled periods of several frames, and (c) the \texttt{N-ray} family coarsens the observation by casting fewer rays. (d) The appearance target locates each disc with categorical bins, sets of positions that produce the same observation. See \cref{sec:rayworld_experiments}.}
\label{fig:rayworld_and_variants}
\end{figure}

Rayworld is a 2D world seen through a single fan of rays. An observer at the origin looks into a trapezoidal frustum in which two discs move at constant velocity for 40 frames. Trajectories in which a disc would hit a wall or the other disc are rejected, so the dynamics are exactly linear. Each frame is rendered into a 1D observation by casting 128 rays, spaced to cross the far edge of the frustum at equal intervals. A ray returns the reflectivity of the nearest disc it hits ($0.4$ or $0.8$ and fixed per disc) or $0$ if it hits neither, giving one noiseless pixel per ray. Depth is never observed directly. It enters only through the number of rays a disc covers, which falls with distance as the rays spread out, and through occlusion.

The Rayworld state $s_t \in \mathbb{R}^8$ holds the position and velocity of both discs and evolves as $s_{t+1} = A\,s_t$, where $A$ advances each position by its velocity, and the frame $o_t = h(s_t) \in [0,1]^R$ over $R$ rays is given by the rendering function $h$. The best possible prediction of the next frame averages $h(A\,s_t)$ over all possible states, $s_t$, which fit the current observations. Nothing here forces a predictor to represent the state itself, and whether it does is what the probes and editors investigate. We probe the state in the Cartesian coordinates $A$ acts in. An edit teleports a single disc, moving it to a random position inside the frustum that is clear of the walls and of the other disc, with both velocities left unchanged. Unlike an Othello edit, every edited state is a valid configuration of the world, with both discs inside the frustum and apart.

The state description we probe for can reveal what implementation the model has learned to achieve its task. To test this, we give Rayworld's continuous positions a categorical counterpart, matching the form of Othello's board. One option is to split the frustum into a uniform grid, with every bin labeled empty, bright disc, or dim disc, depending on which disc's center it contains. Alternatively, we let the observations draw the bins. In an observation, each disc lights a group of adjacent rays, described by its width and lateral position rather than its 2D coordinates. A small move changes this appearance only when the disc crosses into a different ray, so we treat all positions that produce the same appearance as one bin, and call the result the appearance target (\cref{fig:rayworld_and_variants}). An edit then moves a disc between bins, as an Othello edit changes one cell. This places the target in observation space rather than in positions. To keep the number of classes small, the target contains width and lateral position separately, rather than making each combination a separate class. \Cref{app:rw_grid_ablation} compares it with the uniform grid, and we return to it in the discussion.

\paragraph{Rayworld Variants.}
We introduce two variants of Rayworld, each modifying what the observation reveals. In \texttt{blink}, a disc vanishes at random for 1 to 12 frames, with the frame before the blackout and its last hidden frame marked on an edge ray, so its position must be carried across the gap rather than read off the current observation. The \texttt{N-ray} family coarsens the observation by casting fewer rays, for $N \in \set{5, 8, 16, 128}$, to examine how the level of detail affects what a probe can find. The disc radius is doubled so a disc always covers at least one ray. We include \texttt{128-ray}, which matches \texttt{standard}'s ray count but uses the larger discs, to check that disc size alone does not explain our results. \Cref{fig:rayworld_and_variants} illustrates the setup and variants.

\subsection{Model, Training, and Probes}
Every model is the same eight-block transformer of width 512 with about 25M parameters \citep{vaswani2017attention}, differing between environments only in its input and output layers. Rayworld uses linear input and output layers with a squared error loss, and Othello uses a token embedding and a softmax over moves with a cross entropy loss. To control for this difference in interface, we also train an \texttt{8-ray} Rayworld model with Othello's interface, reading and predicting whole frames as tokens (\cref{app:tokenized_rw}). Each model trains for 780,000 steps at batch size 256 on 18M sequences, with another 2M used for validation. The main results tables come from one seed per variant, and retraining each variant from three seeds at a shorter budget of 512,000 steps shows limited variation. We report the spread in \cref{app:metric_spread}.

A decoding probe reads the simulator state out of the model's latent state. The latent state $z_t$ comprises the residual stream, the vector every block adds to and passes on, at every point \citep{elhage2021mathematical}. Its value after block $k$ is residual point $k$, with point 0 as the embedding. \rev{At width 512, $z_t$ thus holds nine 512-dimensional vectors, one per point.} At every point we fit a linear probe, as a single affine map \citep{alain2017probes}, and an MLP probe with one hidden layer of 128 units, the shape used by \citet{li2023othello}. Both are fit on about 1.2M pairs of latent state and simulator state, with inputs standardized. Regression targets are fit by least squares and categorical targets by cross entropy. The targets are the mine/theirs board in Othello, the eight-dimensional state in Rayworld, and the appearance target of \cref{sec:rayworld_experiments}, whose probes use a larger corpus of 200,000 sequences to match their larger output.

\subsection{Editors}

The editors differ only in how they produce $z_t^*$. With $f_{\mathrm{Lin}}(z) = Wz + b$ the linear probe, $W^{+}$ the pseudoinverse of $W$, and $f_{\mathrm{MLP}}$ the MLP probe at the edited residual point,
\begin{align}
\text{PI:}\quad & z_t^* = z_t + \alpha\, W^{+}\big(s_t^* - f_{\mathrm{Lin}}(z_t)\big), \\
\text{GS:}\quad & z_t \leftarrow z_t - \eta\, \nabla_{z_t}\, \mathcal{L}\big(f_{\mathrm{MLP}}(z_t),\, s_t^*\big), \\
\text{IM:}\quad & z_t^* = g(s_t^*).
\end{align}

The first two editors steer $z_t$ through the same probes that decode the state, so their success or failure indicates what those probes have found. Pseudoinverse injection (PI) makes the smallest change to the standardized latent state that moves the linear probe's reading onto the target, scaled by $\alpha$ so that $\alpha = 1$ \rev{matches it} exactly, and writes once at a single residual point. Gradient steering (GS) instead takes 100 gradient steps of size $\eta$ on the latent state, minimizing the MLP probe's loss $\mathcal{L}$ toward the target while weighting entries that should not change by $0.2$. Following \citet{li2023othello}, it writes at a starting point and again at every later point.

The inverse map (IM) runs the probe in reverse, testing whether the model can act as a simulator even when steering fails. We fit $g$ from simulator state to latent state, an MLP of the same size and training setup as the forward probe. At edit time, we overwrite the latent state with $g(s_t^*)$ at one residual point. To rule out memorization, we compare against overwriting with the mean latent state of the ten nearest training states, which IM beats on every Rayworld variant, while in Othello the nearest neighbors never successfully edit (\cref{app:inverse_nearest_neighbor}).

Each variant is scored on 1000 edit cases, each a held-out sequence with one variable changed at step 20. We skip cases whose edit leaves the \rev{legal moves unchanged in Othello, or changes fewer than two rays in Rayworld,} and sweep each editor over residual points. For PI and GS we also sweep the step size ($\alpha$ and $\eta$, respectively).

\subsection{Metrics}
We introduce three metrics, one for decodability and two for editability, each defined relative to a reference so that environments with different outputs can be compared directly.

We measure decodability with Probe Skill $= 1 - L_{\mathrm{probe}} / L_{\mathrm{trivial}}$, a baseline-relative skill score in the sense of \citet{murphy1988skill}, where $L_{\mathrm{probe}}$ is the probe's held-out loss and $L_{\mathrm{trivial}}$ that of a predictor that ignores the model. For continuous targets we use squared error with the training mean as the trivial predictor, so Probe Skill reduces to (out-of-sample) $R^2$. For categorical targets we use the error rate, with the most common class of each cell as the trivial predictor. We also fit the same probes to a randomly initialized model \citep{li2023othello} and to the raw observation sequences, separating what training put into the model from what a probe could extract regardless.

For editability we introduce two measurements, since an edit can implement the target change while degrading the rest of the output, or improve the output without implementing the change. Edit Index measures whether the edit moves the prediction in the right direction, measured only on the part of the output the edit should change: the rays whose values differ between the two worlds in Rayworld, and the cells whose legality the flip changes in Othello. It runs from $-1$, matching the unedited observation, to $+1$, matching the edited one. Edit Fidelity measures whether the output as a whole remains accurate, relative to the unedited prediction: $0$ is no better than making no edit, negative values are worse, and $1$ reproduces the edited observation exactly. With $\operatorname{RMSE}_{\Delta}$ the error over the edit's support,
\begin{equation}
\label{eq:editability_formulae}
\begin{gathered}
d_{\mathrm{edit}} = \operatorname{RMSE}_{\Delta}(\hat{o}_{t+1}^*, o_{t+1}^*), \qquad
d_{\mathrm{uned}} = \operatorname{RMSE}_{\Delta}(\hat{o}_{t+1}^*, o_{t+1}), \\[0.65em]
\text{Edit Index} = \frac{d_{\mathrm{uned}} - d_{\mathrm{edit}}}{d_{\mathrm{uned}} + d_{\mathrm{edit}}}, \qquad
\text{Edit Fidelity} = 1 - \frac{\operatorname{RMSE}(\hat{o}_{t+1}^*, o_{t+1}^*)}{\operatorname{RMSE}(\hat{o}_{t+1}, o_{t+1}^*)}.
\end{gathered}
\end{equation}

We report each editor at its highest Edit Index among settings with Edit Fidelity of at least $0$, falling back to the highest Edit Fidelity where none qualifies. Selecting on Edit Fidelity instead leads to the same conclusions (\cref{app:edit_fidelity_choice}). \rev{Choosing settings on held-out validation edits leaves these conclusions unchanged (\cref{app:robustness}). We say an editor succeeds when its Edit Index reaches $+0.4$ and its Edit Fidelity reaches $0.25$. Its prediction on the changed rays or cells is then over twice as close to the edited observation as to the unedited one, and its error over the whole output falls by at least a quarter.}

\section{Results}
\label{sec:results}

\begin{table}[t]
\centering
\small
\setlength{\tabcolsep}{3pt}

\begin{tabular}{ll PP g E g EF g EF g EF}
\hline
\noalign{\vspace{0.25em}}
& & \multicolumn{2}{c}{Probe Skill} & & \multicolumn{1}{c}{} & & \multicolumn{8}{c}{Editor} \\
\noalign{\vspace{0.25em}}
\cline{3-4} \cline{8-15}
\noalign{\vspace{0.25em}}
& & \multicolumn{1}{c}{Obs.} & \multicolumn{1}{c}{Trained} & & \multicolumn{1}{c}{Unedited} & & \multicolumn{2}{c}{PI} & & \multicolumn{2}{c}{GS} & & \multicolumn{2}{c}{IM} \\
Class & Variant & \multicolumn{1}{c}{Lin.} & \multicolumn{1}{c}{Lin.} & & \multicolumn{1}{c}{Index} & &
\multicolumn{1}{c}{Index} & \multicolumn{1}{c}{Fid.} & &
\multicolumn{1}{c}{Index} & \multicolumn{1}{c}{Fid.} & &
\multicolumn{1}{c}{Index} & \multicolumn{1}{c}{Fid.} \\
\hline
\noalign{\vspace{0.1em}}
\multirow{4}{*}{\rotatebox[origin=c]{90}{Othello}}
& \texttt{standard}        & 0.792 & 0.975 & & -0.93 & & +0.82 & 0.70 & & +0.83 & 0.72 & & +0.81 & 0.62 \\
& \texttt{adjacent-flip}   & 0.888 & 0.947 & & -0.96 & & +0.35\dg & 0.17 & & -0.06 & 0.21 & & +0.66 & 0.49\dg \\
& \texttt{adjacent-noflip} & 0.988 & 0.988 & & -0.96 & & -0.23\dg & 0.19 & & -0.96 & -0.01 & & -0.03\dg & 0.26 \\
& \texttt{standard-noflip} & 1.000 & 1.000 & & -0.98 & & -0.95 & 0.00 & & -0.97 & -0.01 & & -0.93 & -0.01 \\
\cline{1-15}
\noalign{\vspace{0.1em}}
\multirow{6}{*}{\rotatebox[origin=c]{90}{\shortstack[c]{Rayworld\\(continuous)}}}
& \texttt{standard} (128 rays) & 0.305 & 0.872 & & -0.93 & & -0.10 & 0.04 & & -0.16 & 0.04 & & +0.59 & 0.66 \\
& \texttt{blink} (128 rays)    & 0.216 & 0.907 & & -0.92 & & +0.02 & 0.01 & & -0.09 & 0.02 & & +0.51 & 0.55 \\
& \texttt{128-ray} (big discs) & 0.377 & 0.914 & & -0.94 & & -0.02 & 0.02 & & -0.10 & 0.03 & & +0.57 & 0.68 \\
& \texttt{16-ray}   & 0.360 & 0.892 & & -0.91 & & +0.16 & 0.04 & & -0.10 & 0.10 & & +0.66 & 0.71 \\
& \texttt{8-ray}    & 0.329 & 0.886 & & -0.90 & & +0.21 & 0.00 & & -0.07 & 0.12 & & +0.71 & 0.73 \\
& \texttt{5-ray}    & 0.277 & 0.850 & & -0.89 & & +0.14 & 0.15 & & -0.10 & 0.14 & & +0.81 & 0.77 \\
\cline{1-15}
\noalign{\vspace{0.1em}}
\multirow{4}{*}{\rotatebox[origin=c]{90}{\shortstack[c]{Rayworld\\(categorical)}}}
& \texttt{128-ray} (3736 bins) & 0.077 & 0.261 & & -0.94 & & -0.31 & 0.02 & & +0.31 & 0.21 & & +0.64 & 0.71 \\
& \texttt{16-ray} (77 bins)    & 0.356 & 0.879 & & -0.91 & & -0.06 & 0.04 & & +0.28 & 0.15 & & +0.82 & 0.71 \\
& \texttt{8-ray} (30 bins)     & 0.477 & 0.935 & & -0.90 & & +0.38 & 0.05 & & +0.46 & 0.46 & & +0.87 & 0.73 \\
& \texttt{5-ray} (14 bins)     & 0.522 & 0.926 & & -0.91 & & +0.51 & 0.30 & & +0.60 & 0.54 & & +0.91 & 0.75 \\
\noalign{\vspace{0.75em}}
\hline
\end{tabular}
\caption{\rev{Decodability and editability} (\textcolor{heatHiTxt}{higher} is better, \textcolor{heatLo}{lower} is worse). \rev{Probe Skill measures decodability, shown for the linear probes on the raw observation history baselines (Obs.) and on the trained models (Trained), with full results in \cref{tab:decodability}.} For each variant and editor, we report the highest Edit Index across residual points and step sizes among settings with Edit Fidelity $\geq 0$, with the Edit Fidelity at that setting. Where no setting reaches $0$ we select for the highest Edit Fidelity. Scores are averaged over 1000 edit cases. Edit Index compares the edited prediction against the edited and unedited ground truth over only the rays or cells that change. Edit Fidelity measures whole-output improvement over the unedited prediction, from $1$ for a perfect match to $0$ for no improvement. Cells marked $\dagger$ have a seed standard deviation above $0.1$ (\cref{app:metric_spread}). See \cref{sec:results}.}
\label{tab:editability}
\end{table}

\rev{\Cref{tab:editability} reports} decodability and editability for every variant, from one training run each. Retraining each variant from two further seeds gives a seed standard deviation of at most $0.004$ for every trained probe's Probe Skill, and below $0.04$ for the Edit Index of every editor that \rev{succeeds}. Cells whose spread exceeds $0.1$ carry a dagger, and all but IM's fidelity on \texttt{adjacent-flip} are failing editors whose selected setting flips between seeds at the fidelity cutoff (\cref{app:metric_spread}). Every model predicts its environment well before any probing or editing begins (\cref{app:predict_skill}).

\paragraph{Decodability.}
The simulator state is decodable in every variant, with the MLP probe reaching a Probe Skill of at least $0.88$ and the linear probe close behind, so what separates the variants is the baselines. In Othello, training adds something the baselines lack only when flipping is on. Once flipping is removed the board can be read off the move sequence itself, so the \texttt{adjacent-noflip} model only matches its observation baseline, and every probe on \texttt{standard-noflip} scores a perfect $1.000$, even on a randomly initialized model. In Rayworld, \rev{the frames themselves are far from linearly decodable (at most $0.38$), but} a randomly initialized model already comes within $0.03$ of the trained one under the MLP probe on every variant except \texttt{blink}, where a hidden disc must be carried forward rather than located in the current frame. \rev{Only the 3736-bin categorical target on \texttt{128-ray} is decoded poorly ($0.26$, against $0.08$ from the frames).}

\paragraph{Editability.}
Every editor edits \texttt{standard} Othello, consistent with \citet{li2023othello} and \citet{nanda2023linear}, each carrying the predicted move distribution most of the way to the flipped board. Every change to the rules breaks this. With flipping off, no editor \rev{succeeds} on either no-flip variant, even on \texttt{adjacent-noflip}, where legal moves still depend on token color. \rev{In} \texttt{adjacent-flip} the editors separate\rev{, as the inverse map still succeeds} while PI and GS reach at most $+0.35$ at low fidelity. 

\begin{figure}[t]
\centering
\includegraphics[width=0.9\linewidth]{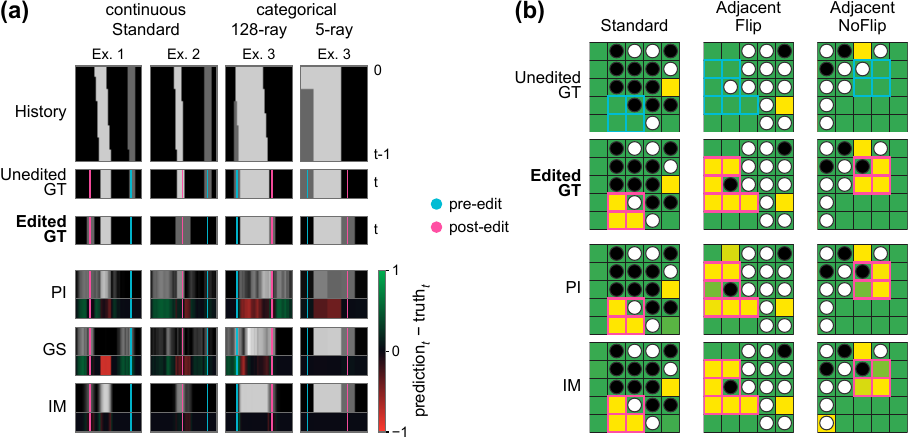}
\caption{Qualitative examples of edited predictions in (a) Rayworld and (b) Othello, \rev{for a subset of} environment variants and editors. \rev{\Cref{app:additional_edit_visuals} shows every variant and editor.} Quantitative results in \cref{tab:editability}. Cyan and magenta coloring highlight the disc or cells that changed as a result of the edit.}
\label{fig:qualitative_edits}
\end{figure}

In Rayworld this separation \rev{holds across all the continuous variants:} the probe-derived editors fail while \rev{the inverse map succeeds each time}, including three further variants in \cref{app:additional_rayworld_runs}. PI's write reaches its target in probe space, since at $\alpha = 1$ the linear probe reads the edited state exactly, yet the predicted frame does not move. A latent state that renders the teleported disc therefore exists and can be written, but not along any direction the probes provide. Discretizing the target changes this. It raises GS on the \texttt{128-ray} model, and as the rays coarsen both probe-derived editors rise steadily, to $+0.51$ and $+0.60$ at 5 rays. Since the number of bins falls with the ray count, this trend cannot separate coarseness from target size. At 8 rays, evenly spaced bins that ignore the observation also improve on continuous positions, but the appearance bins improve most (\cref{app:rw_grid_ablation}). \Cref{fig:qualitative_edits} shows representative edits qualitatively.

\begin{figure}[t]
\centering
\includegraphics[width=1.0\linewidth]{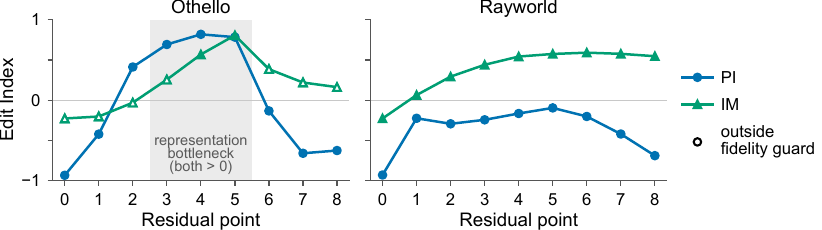}
\caption{\rev{Edit Index of PI and the inverse map at every residual point of the \texttt{standard} Othello (left) and Rayworld (right) models. The shaded points mark Othello's bottleneck. Rayworld has no such bottleneck, and the inverse map succeeds once the model has partly rendered the next frame (\cref{app:im_editability_trends,app:inverse_controls}). GS is not shown, since it writes at every point from where it starts onward.}}
\label{fig:editability_main}
\end{figure}

\begin{revblock}
\paragraph{Where edits succeed.}
Investigating where edits succeed across a network's layers can isolate where representations form and how the model is computing its predictions (\cref{fig:editability_main}). The Othello model appears to compress its history into a board at points 4 and 5 and predict from it. PI and the inverse map peak there and fail from point 6 onward, and GS succeeds only when it starts before point 6. In earlier blocks the model is still gathering information from previous steps, and in later blocks it computes the legal moves without looking back at the history (\cref{app:im_editability_trends}). The Rayworld model instead appears to render the next frame gradually. PI fails at every point, while the inverse map succeeds from point 3 and plateaus around point 5. The cause is revealed by passing the latent state at each point through the decoder (\cref{app:inverse_controls}). A rough version of the next frame appears by point 3 and most of it by point 5, so the inverse map succeeds by writing this partial rendering.
\end{revblock}

\section{Discussion}
\begin{revblock}
In Othello, the variants combine two rule changes, and each makes the game simpler to play (\cref{sec:othello_experiments}). Each change also lowers editability. Under the standard rules, every editor succeeds on reachable and unreachable boards alike, with one probe or map fit across all time steps. With adjacent placement, only the inverse map succeeds, and it edits unreachable boards less well, at $+0.61$ against $+0.78$ for reachable ones (\cref{app:legal_vs_illegal_othello}).

Without token flipping, a single probe still decodes the board at every step. On \texttt{adjacent-noflip}, the model can be edited under the right circumstances, but not robustly. Probe steering fails, and the inverse map edits single flips only when it is fit at a single time step, at an Edit Index of $+0.86$ (\cref{app:robustness}). No game could reach these boards, so reachability is not the barrier. On \texttt{standard-noflip}, each cell determines its color, so the data never reveals how the rules would treat a token switched to the other color. Prediction never requires the model to represent color apart from the cell, which makes this variant the limiting case of a minimal sufficient statistic coarser than the simulator's state (\cref{sec:forcing_state}), and every edit fails.

In Rayworld, the probes fail to find a direction that steers continuous positions in any variant of our fixed architecture, including with Othello's token interface, along encoding directions, or when written at every later point (\cref{app:additional_rayworld_runs,app:tokenized_rw,app:steering_variants}). Our hypothesis is that the early layers shift the recent frames in observation space, bypassing any 2D state, and the later layers refine this rendering of the next frame. The inverse map can drive the model from a simulator it may never have learned, but what it writes is the model's own rendering rather than the simulator's state.
\end{revblock}

The inverse map's fit and its edits also \rev{diverge} in Rayworld, where it explains far less of the latent variance ($R^2$ of $0.32$ to $0.58$ in \rev{\cref{tab:decodability}}) yet \rev{succeeds on every Rayworld variant. On the 128-ray variants its $R^2$ stays below $0.4$ even when fit at the edit step alone (\cref{app:robustness}), so} much of that variance appears to have little bearing on the prediction.

\begin{revblock}
\subsection{Forcing a Simulator-Like State}
\label{sec:forcing_state}
To predict well, a model must keep some summary of its history. Computational mechanics calls the smallest summary that does not worsen predictive performance the causal state \citep{shalizi2001computational}. Let $x_{\le t}$ be the history of observations and inputs up to step $t$, $o_{>t}$ the future observations, and $u_{>t}$ the future inputs, which are set from outside, not predicted, such as a camera's viewpoint. Two histories $x_{\le t}$ and $x'_{\le t}$ are grouped into one causal state when they predict the same future under every future input \citep{barnett2015computational},
\begin{equation}
\label{eq:causal_state}
x_{\le t} \sim x'_{\le t} \iff P(o_{>t} \mid x_{\le t}, u_{>t}) = P(o_{>t} \mid x'_{\le t}, u_{>t}) \;\text{ for all } u_{>t}.
\end{equation}
The causal state is the minimal sufficient statistic of the history, and is all a model must learn to predict optimally. Our simulators generate the observations, so their state is at least a sufficient statistic and the causal state can only be coarser than it. Refining the causal state is therefore guaranteed to force the model to represent more of the simulator's state. In Othello, we hypothesize that more complex rules leave fewer representations simpler than the board that predict equally well, so the causal state is closer to the board. In Rayworld, we hypothesize that the causal state is coarser than the simulator's state. We propose this is an effective definition of what it means to take a shortcut. Since the model already predicts nearly optimally (\cref{app:predict_skill}), more scale gives it no reason to represent distinctions the causal state discards.

\Cref{eq:causal_state} also shows how to refine the causal state. The more future inputs a process allows, the more futures two histories must predict alike to be grouped together (\cref{app:pose_delay}). We test this by rendering each Rayworld frame from a different viewpoint. The model must commit to a latent state before it knows the next viewpoint, and render the next frame from that state alone, so that state must serve every viewpoint. On \texttt{8-ray}, this raises the Edit Index of PI and GS on continuous positions from $+0.21$ and $-0.07$ to $+0.55$ and $+0.42$ (\cref{tab:pose_delay}), with a smaller gain on \texttt{128-ray}. When the rendering step can instead read the whole history, the gain disappears, as the theory predicts.
\end{revblock}

\section{Conclusion}
We investigated when a generative model can be driven as a simulator by setting its latent state, and found that the answer depends on the environment in ways decodability does not predict. In Othello, \rev{each simplification of the rules lowers editability.} In Rayworld, probe-derived editors cannot steer the continuous state, yet an inverse map sets the model from it anyway, \rev{by approximating the model's partial rendering of the next frame. We relate both outcomes to whether prediction pressures a model to compress its history into the simulator's state, and use this framing to suggest a fix.} Our editability concerns a single step, whereas a simulator also carries its state forward under consistent rules (see \cref{app:limitations} for further limitations). Rewriting the history lets an edit persist through a rollout (\cref{app:history_editing}). \rev{We plan to extend these ideas toward a general theory of how to build a simulator into a generative model, to achieve open-ended virtual worlds with persistent and controllable state.}

\newpage
\section{Reproducibility Statement}
\label{sec:reproducibility}
We release the code for every environment, model, probe, editor, and metric at \url{https://github.com/SevanBrodjian/generative-models-as-simulators}. The trained weights of every model, including the seed replicates, and the evaluation and edit datasets of every variant are at \url{https://huggingface.co/datasets/AnonymousCompute/generative-models-as-simulators}. The training corpora are not hosted, but the released scripts regenerate each one from its recorded seeds. \Cref{sec:experimental_setup} describes the setup, \cref{app:implementation} lists the remaining settings, and \cref{app:metric_spread} reports the spread over training seeds and probe refits.

\bibliography{iclr2027_conference}
\bibliographystyle{iclr2027_conference}

\newpage

\appendix

\section{Appendix}

\noindent\textbf{Contents}
\medskip

\begingroup
\newcommand{\appline}[1]{\noindent\makebox[2.5em][l]{\ref{#1}}\nameref{#1}\dotfill\pageref{#1}\par}
\appline{app:limitations}
\appline{app:predict_skill}
\appline{app:metric_spread}
\appline{app:legal_vs_illegal_othello}
\appline{app:tokenized_rw}
\appline{app:additional_rayworld_runs}
\appline{app:im_editability_trends}
\appline{app:rw_grid_ablation}
\appline{app:inverse_nearest_neighbor}
\appline{app:history_editing}
\appline{app:edit_fidelity_choice}
\appline{app:implementation}
\appline{app:additional_edit_visuals}
\appline{app:inverse_controls}
\appline{app:steering_variants}
\appline{app:robustness}
\appline{app:pose_delay}
\endgroup
\bigskip

\subsection{Limitations}
\label{app:limitations}
Our editability is a statement about only one step. A simulator also carries its state forward under consistent rules, and we test only whether a single write changes the next prediction as it should. Rewriting the history so that an edit persists through a rollout is one route, which we demonstrate in \cref{app:history_editing}, and extending the metrics to rollouts is the natural next step. Every model here is one transformer at one depth, so some of what we find may be a property of that architecture or that depth rather than of the environments, and recurrent or diffusion models may behave differently. The environments also differ in difficulty in ways we cannot fully control, and a harder environment may be less editable for reasons unrelated to how its state is represented.

\FloatBarrier
\subsection{Predictive Loss \rev{and Decodability} of Trained Models}
\label{app:predict_skill}

\begin{figure}[ht]
\centering
\includegraphics[width=0.89\linewidth]{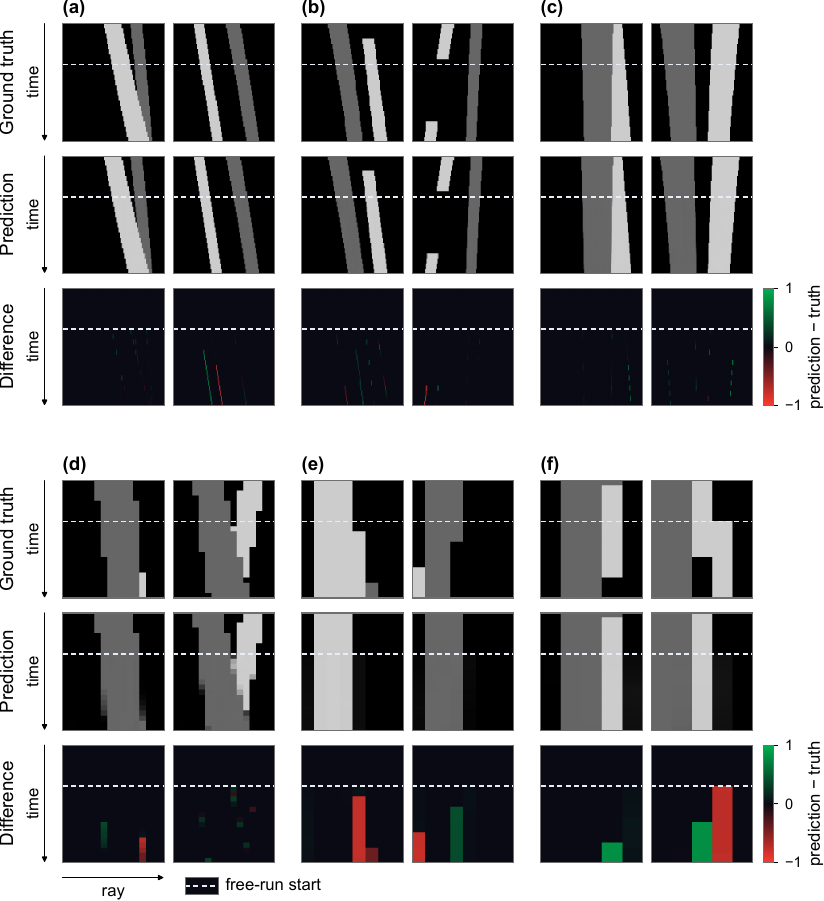}
\caption{Randomly sampled demonstrations of Rayworld predictive accuracy in (a) \texttt{standard}, (b) \texttt{blink}, (c) \texttt{128-ray}, (d) \texttt{16-ray}, (e) \texttt{8-ray}, and (f) \texttt{5-ray}.}
\label{fig:rayworld_predictions}
\end{figure}

\begin{figure}[ht]
\centering
\includegraphics[width=1.0\linewidth]{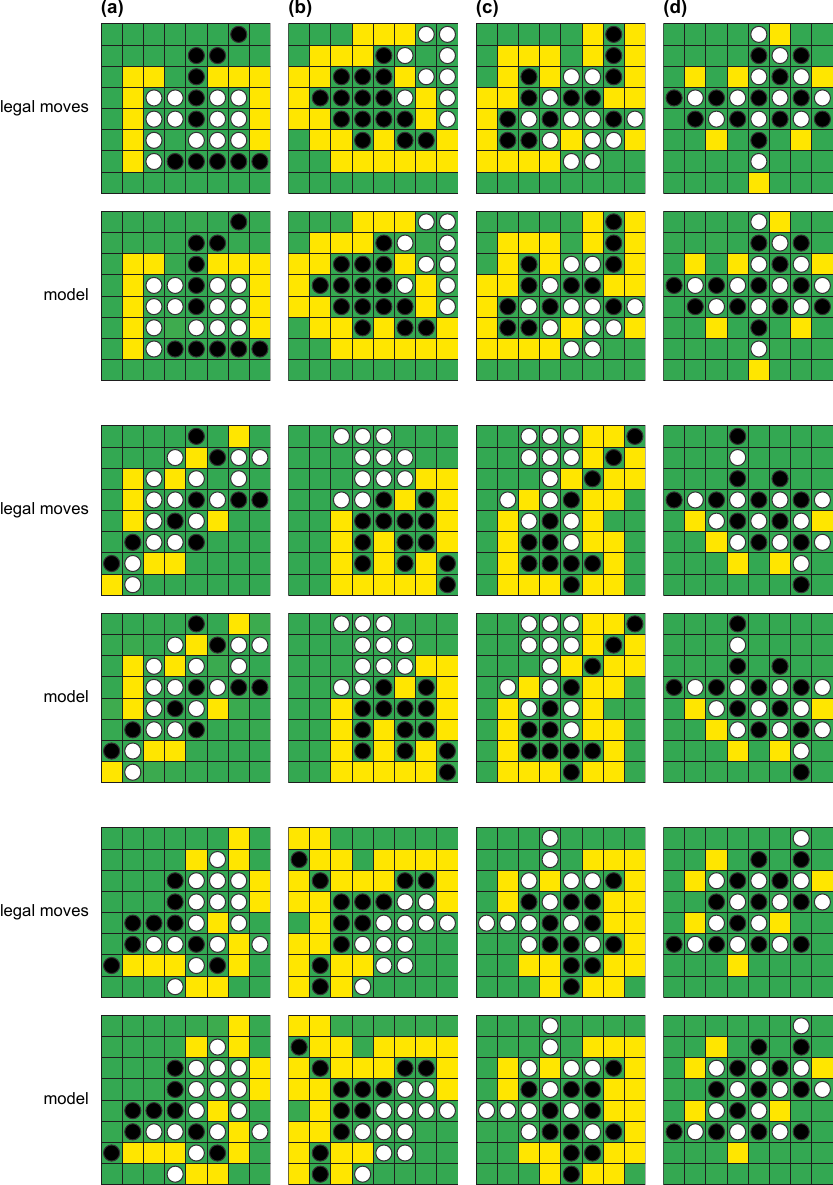}
\caption{Randomly sampled demonstrations of Othello predictive accuracy in (a) \texttt{standard}, (b) \texttt{adjacent-flip}, (c) \texttt{adjacent-noflip}, and (d) \texttt{standard-noflip}.}
\label{fig:othello_predictions}
\end{figure}

Every model predicts its environment nearly as well as the environment allows, so none of our editing results can be blamed on a weak predictor. \Cref{tab:predictive_skill} places each model's held-out loss between two references. The trivial predictor ignores the history and always makes the same guess, the overall frequency of each move in Othello and the average frame in Rayworld. The Bayes floor is the loss of the best possible predictor, which knows the environment's rules and is limited only by what the history leaves uncertain. The Othello floor is exact, since the generator draws each move uniformly from the legal moves. In Rayworld, a sequence is fixed by the discs' starting positions and velocities, but different starting states can light the same rays, so the frames seen so far do not pin down the next one. The best predictor averages the next frame over every starting state consistent with those frames, weighted by how likely the generator was to draw it. We approximate this average with sampled starting states, dropping each one as soon as it fails to reproduce an observed frame. The floor rises as the rays coarsen and leave more of the state undetermined. Every model closes at least 99\% of the gap from the trivial predictor to its floor. \Cref{fig:rayworld_predictions,fig:othello_predictions} show this accuracy directly. On \texttt{5-ray} a rollout can keep a disc on the wrong ray for several steps. In Othello, boards shaded by the model's predicted distribution are nearly indistinguishable from boards shaded by the true legal moves. \rev{\Cref{tab:decodability} gives the decodability results summarized in \cref{tab:editability}, including the MLP probes, the randomly initialized model, and the inverse map's fit to the latent state.}

\begin{table}[ht]
\centering
\small
\setlength{\tabcolsep}{4pt}
\begin{tabular}{ll PP G PP G PP G PP}
\hline
\noalign{\vspace{0.25em}}
& & \multicolumn{11}{c}{Probe Type} \\
\noalign{\vspace{0.25em}}
\cline{3-13}
\noalign{\vspace{0.25em}}
& & \multicolumn{2}{c}{Observation} & & \multicolumn{2}{c}{Random init} & & \multicolumn{2}{c}{Trained} & & \multicolumn{2}{c}{Inverse Map MLP} \\
Class & Variant
& \multicolumn{1}{c}{Lin.} & \multicolumn{1}{c}{MLP} & &
\multicolumn{1}{c}{Lin.} & \multicolumn{1}{c}{MLP} & &
\multicolumn{1}{c}{Lin.} & \multicolumn{1}{c}{MLP} & &
\multicolumn{1}{c}{Best Point} & \multicolumn{1}{c}{Edit Point} \\
\hline
\noalign{\vspace{0.1em}}
\multirow{4}{*}{\rotatebox[origin=c]{90}{Othello}}
& \texttt{standard}        & 0.792 & 0.809 & & 0.577 & 0.583 & & 0.975 & 0.976 & & 0.884 & 0.827 \\
& \texttt{adjacent-flip}   & 0.888 & 0.914 & & 0.674 & 0.700 & & 0.947 & 0.970 & & 0.968 & 0.959 \\
& \texttt{adjacent-noflip} & 0.988 & 0.983 & & 0.581 & 0.588 & & 0.988 & 0.990 & & 0.978 & 0.855 \\
& \texttt{standard-noflip} & 1.000 & 1.000 & & 1.000 & 1.000 & & 1.000 & 1.000 & & 0.947 & 0.855 \\
\cline{1-13}
\noalign{\vspace{0.1em}}
\multirow{6}{*}{\rotatebox[origin=c]{90}{Rayworld}}
& \texttt{standard} (128 rays) & 0.305 & 0.885 & & 0.738 & 0.945 & & 0.872 & 0.973 & & 0.741 & 0.339 \\
& \texttt{blink} (128 rays)    & 0.216 & 0.809 & & 0.608 & 0.870 & & 0.907 & 0.979 & & 0.613 & 0.338 \\
& \texttt{128-ray} (big discs) & 0.377 & 0.919 & & 0.859 & 0.970 & & 0.914 & 0.981 & & 0.652 & 0.321 \\
& \texttt{16-ray}   & 0.360 & 0.907 & & 0.890 & 0.949 & & 0.892 & 0.965 & & 0.554 & 0.516 \\
& \texttt{8-ray}    & 0.329 & 0.877 & & 0.883 & 0.919 & & 0.886 & 0.932 & & 0.585 & 0.580 \\
& \texttt{5-ray}    & 0.277 & 0.829 & & 0.855 & 0.871 & & 0.850 & 0.883 & & 0.617 & 0.469 \\
\hline
\end{tabular}
\caption{Decodability (\textcolor{heatHiTxt}{higher} is better, \textcolor{heatLo}{lower} is worse), measured as Probe Skill at the trained model's best residual point, alongside random-initialization and observation baselines fit with the same probes on the same sequences. The inverse map columns give its $R^2$ in predicting latent state from simulator state, at its best residual point and at the point where it edits best. See \cref{sec:results}.}
\label{tab:decodability}
\end{table}

\begin{table}[ht]
\centering
\small
\setlength{\tabcolsep}{4pt}
\begin{tabular}{ll rrrr}
\hline
\noalign{\vspace{0.25em}}
Class & Variant & \multicolumn{1}{c}{Trivial} & \multicolumn{1}{c}{Bayes floor} & \multicolumn{1}{c}{Model} & \multicolumn{1}{c}{Excess} \\
\hline
\noalign{\vspace{0.1em}}
\multirow{4}{*}{\shortstack[l]{Othello\\(nats per move)}}
& \texttt{standard}        & 4.094 & 2.011 & 2.029 & 0.018 \\
& \texttt{adjacent-flip}   & 4.094 & 2.295 & 2.301 & 0.005 \\
& \texttt{adjacent-noflip} & 4.094 & 2.433 & 2.435 & 0.002 \\
& \texttt{standard-noflip} & 4.094 & 1.679 & 1.680 & 0.001 \\
\cline{1-6}
\multirow{6}{*}{\shortstack[l]{Rayworld\\(MSE $\times 10^{-3}$)}}
& \texttt{standard} (128 rays) & 72.6  & 0.80 $\pm$ 0.04 & 1.06 & 0.26 $\pm$ 0.04 \\
& \texttt{blink} (128 rays)    & 66.0  & 0.98 $\pm$ 0.03 & 1.26 & 0.28 $\pm$ 0.03 \\
& \texttt{128-ray}  & 92.4  & 0.75 $\pm$ 0.03 & 0.94 & 0.19 $\pm$ 0.03 \\
& \texttt{16-ray}   & 97.2  & 3.76 $\pm$ 0.13 & 4.00 & 0.23 $\pm$ 0.13 \\
& \texttt{8-ray}    & 100.7 & 5.62 $\pm$ 0.14 & 5.74 & 0.12 $\pm$ 0.14 \\
& \texttt{5-ray}    & 105.5 & 6.86 $\pm$ 0.18 & 7.05 & 0.19 $\pm$ 0.18 \\
\cline{1-6}
\multirow{2}{*}{\shortstack[l]{Rayworld\\(\texttt{8-ray} tokens)}}
& next-token CE (nats per frame)   & 5.176 & 0.441 $\pm$ 0.011 & 0.464 & 0.023 $\pm$ 0.011 \\
& mean frame (MSE $\times 10^{-3}$) & 100.7 & 5.62 $\pm$ 0.14 & 5.73  & 0.11 $\pm$ 0.14 \\
\hline
\end{tabular}
\caption{Held-out predictive loss of every trained model against the Bayes floor of its environment and a trivial history-blind predictor. Details in \cref{app:predict_skill}.}
\label{tab:predictive_skill}
\end{table}

\FloatBarrier
\subsection{Metric Variance and Confidence Intervals}
\label{app:metric_spread}

The seed spread is small compared to the differences between variants. We retrain every variant from two further seeds and compare the three members of each set at 512,000 steps, namely the two new models and the main run's own checkpoint at that step, each scored exactly as in the main tables. \Cref{tab:seed_spread} gives the standard deviation over the three seeds.

No Probe Skill varies by more than $0.004$, and the Edit Index of every successful editor varies by at most $0.034$. Each step of IM's increasing editability as Rayworld's rays coarsen is at least six standard deviations. Four cells exceed $0.1$ and carry a dagger in \cref{tab:editability}. Three are failing editors on the two adjacency variants, PI on both and IM on \texttt{adjacent-noflip}, and their spread comes mostly from selecting a different setting on each seed rather than from the models. Holding the main run's setting fixed, the Edit Index of both PI cells varies by only $0.05$ to $0.07$. IM on \texttt{adjacent-noflip} stays unstable even then, at $0.32$, and on one seed no setting passes the cutoff, so it selects for highest fidelity instead and reports a much lower index. We keep the per-seed selection in \cref{tab:seed_spread}, since it is what rerunning the full pipeline would give. The fourth dagger, IM's Edit Fidelity on \texttt{adjacent-flip}, is not due to selection. All three seeds write at the same point with an Edit Index of $0.62 \pm 0.03$, while the fidelity ranges from $0.40$ to $0.59$.

Outside the daggered cells, the seed means sit within $0.05$ of each Edit Index in \cref{tab:editability}, which reports the same runs at 780,000 steps. Edit Fidelity can differ by up to $0.26$ where the selected setting moves between the two checkpoints. PI on \texttt{standard} Othello, for example, is selected at a larger step size at 512,000 steps, with a mean Edit Fidelity of $0.48$ against $0.70$.

Refitting the probes on a fixed model has an even smaller spread. Refitting the linear probe with six to ten seeds, which change its initialization and held-out split, moves Probe Skill by at most $0.0007$ and PI's Edit Index by at most $0.016$ on \texttt{standard} Othello, the three \texttt{adjacent-flip} models, and the categorical target of the \texttt{8-ray} model. The MLP probes implementing GS were not refit.

\begin{table}[ht]
\centering
\small
\setlength{\tabcolsep}{4pt}
\begin{tabular}{ll rr G rr G rr G rr}
\hline
\noalign{\vspace{0.25em}}
& & \multicolumn{2}{c}{Probe Skill} & & \multicolumn{2}{c}{PI} & & \multicolumn{2}{c}{GS} & & \multicolumn{2}{c}{IM} \\
\noalign{\vspace{0.25em}}
\cline{3-4} \cline{6-7} \cline{9-10} \cline{12-13}
\noalign{\vspace{0.25em}}
Class & Variant & \multicolumn{1}{c}{Lin.} & \multicolumn{1}{c}{MLP} & &
\multicolumn{1}{c}{Index} & \multicolumn{1}{c}{Fid.} & &
\multicolumn{1}{c}{Index} & \multicolumn{1}{c}{Fid.} & &
\multicolumn{1}{c}{Index} & \multicolumn{1}{c}{Fid.} \\
\hline
\noalign{\vspace{0.1em}}
\multirow{4}{*}{\rotatebox[origin=c]{90}{Othello}}
& \texttt{standard}        & 0.003 & 0.002 & & 0.015 & 0.054 & & 0.012 & 0.007 & & 0.034 & 0.038 \\
& \texttt{adjacent-flip}   & 0.000 & 0.001 & & 0.245\dg & 0.031 & & 0.073 & 0.038 & & 0.028 & 0.108\dg \\
& \texttt{adjacent-noflip} & 0.001 & 0.001 & & 0.196\dg & 0.085 & & 0.005 & 0.010 & & 0.422\dg & 0.077 \\
& \texttt{standard-noflip} & 0.000 & 0.000 & & 0.001 & 0.001 & & 0.002 & 0.002 & & 0.026 & 0.002 \\
\cline{1-13}
\noalign{\vspace{0.1em}}
\multirow{6}{*}{\rotatebox[origin=c]{90}{\shortstack[c]{Rayworld\\(continuous)}}}
& \texttt{standard} (128 rays) & 0.002 & 0.000 & & 0.042 & 0.010 & & 0.027 & 0.026 & & 0.005 & 0.020 \\
& \texttt{blink} (128 rays)    & 0.002 & 0.001 & & 0.060 & 0.016 & & 0.018 & 0.020 & & 0.012 & 0.007 \\
& \texttt{128-ray} (big discs) & 0.004 & 0.000 & & 0.009 & 0.010 & & 0.027 & 0.023 & & 0.006 & 0.008 \\
& \texttt{16-ray}   & 0.002 & 0.000 & & 0.021 & 0.020 & & 0.010 & 0.009 & & 0.007 & 0.004 \\
& \texttt{8-ray}    & 0.002 & 0.000 & & 0.022 & 0.041 & & 0.030 & 0.017 & & 0.006 & 0.008 \\
& \texttt{5-ray}    & 0.002 & 0.000 & & 0.036 & 0.051 & & 0.015 & 0.016 & & 0.009 & 0.007 \\
\cline{1-13}
\noalign{\vspace{0.1em}}
\multirow{4}{*}{\rotatebox[origin=c]{90}{\shortstack[c]{Rayworld\\(categorical)}}}
& \texttt{128-ray} (3736 bins) & 0.001 & 0.001 & & 0.024 & 0.020 & & 0.014 & 0.019 & & 0.003 & 0.005 \\
& \texttt{16-ray} (77 bins)    & 0.002 & 0.000 & & 0.014 & 0.019 & & 0.006 & 0.065 & & 0.002 & 0.004 \\
& \texttt{8-ray} (30 bins)     & 0.000 & 0.000 & & 0.006 & 0.027 & & 0.014 & 0.009 & & 0.005 & 0.001 \\
& \texttt{5-ray} (14 bins)     & 0.000 & 0.000 & & 0.017 & 0.029 & & 0.012 & 0.016 & & 0.003 & 0.000 \\
\noalign{\vspace{0.75em}}
\hline
\end{tabular}
\caption{Seed standard deviation of the trained models' Probe Skill and of each editor's Edit Index and Edit Fidelity, over three training seeds per variant at 512,000 steps. Rows and settings as in \cref{tab:decodability,tab:editability}. Cells marked $\dagger$ exceed $0.1$.}
\label{tab:seed_spread}
\end{table}

\FloatBarrier
\subsection{Comparing Reachable vs Unreachable Target States in Othello}
\label{app:legal_vs_illegal_othello}

An Othello edit flips one token of a real board, and nothing guarantees that any game could reach the result. We therefore split the edit cases by whether some legal game ends on exactly the target board, which we decide by an exhaustive search over games (\cref{tab:legal_illegal}). On \texttt{standard} Othello every editor edits both kinds about equally well. On \texttt{adjacent-flip}, IM \rev{succeeds} on both but does better on reachable boards. Neither no-flip variant has any reachable flipped board, so the failures in \cref{tab:editability} there are all on unreachable targets.

To test whether a model that fails on single flips can be edited toward a reachable board, we flip two tokens of opposite color at once, which keeps each color's count (\cref{tab:two_flip}). On \texttt{standard} and \texttt{adjacent-flip} the editors behave as they do for single flips. On \texttt{adjacent-noflip}, IM \rev{succeeds} on the reachable pairs at $+0.88$ with an Edit Fidelity of $0.78$, comparable to \texttt{standard} Othello, while PI and GS fail. On the unreachable pairs IM still moves the changed cells toward the target, at $+0.65$, but an Edit Fidelity of $0.08$ shows it degrades the rest of the prediction. On \texttt{standard-noflip} no two-token edit is reachable, since every token's color there is fixed by its cell, and no editor \rev{succeeds}.

\begin{table}[ht]
\centering
\small
\setlength{\tabcolsep}{4pt}
\begin{tabular}{ll r E G EF G EF G EF}
\hline
\noalign{\vspace{0.25em}}
& & & \multicolumn{1}{c}{} & & \multicolumn{8}{c}{Editor} \\
\noalign{\vspace{0.25em}}
\cline{6-13}
\noalign{\vspace{0.25em}}
& & & \multicolumn{1}{c}{Unedited} & & \multicolumn{2}{c}{PI} & & \multicolumn{2}{c}{GS} & & \multicolumn{2}{c}{IM} \\
Variant & Target & \multicolumn{1}{c}{$n$} & \multicolumn{1}{c}{Index} & &
\multicolumn{1}{c}{Index} & \multicolumn{1}{c}{Fid.} & &
\multicolumn{1}{c}{Index} & \multicolumn{1}{c}{Fid.} & &
\multicolumn{1}{c}{Index} & \multicolumn{1}{c}{Fid.} \\
\hline
\noalign{\vspace{0.1em}}
\multirow{2}{*}{\texttt{standard}}
& legal   & 441 & -0.93 & & +0.79 & 0.70 & & +0.84 & 0.73 & & +0.83 & 0.62 \\
& illegal & 555 & -0.93 & & +0.84 & 0.71 & & +0.82 & 0.71 & & +0.79 & 0.62 \\
\cline{1-13}
\multirow{2}{*}{\texttt{adjacent-flip}}
& legal   & 335 & -0.95 & & +0.37 & 0.27 & & +0.09 & 0.24 & & +0.78 & 0.51 \\
& illegal & 614 & -0.97 & & +0.35 & 0.12 & & -0.12 & 0.19 & & +0.61 & 0.48 \\
\cline{1-13}
\multirow{2}{*}{\texttt{adjacent-noflip}}
& legal   & 0    & \multicolumn{1}{r}{--} & & \multicolumn{1}{r}{--} & \multicolumn{1}{r}{--} & & \multicolumn{1}{r}{--} & \multicolumn{1}{r}{--} & & \multicolumn{1}{r}{--} & \multicolumn{1}{r}{--} \\
& illegal & 1000 & -0.96 & & -0.23 & 0.19 & & -0.96 & -0.01 & & -0.03 & 0.26 \\
\cline{1-13}
\multirow{2}{*}{\texttt{standard-noflip}}
& legal   & 0    & \multicolumn{1}{r}{--} & & \multicolumn{1}{r}{--} & \multicolumn{1}{r}{--} & & \multicolumn{1}{r}{--} & \multicolumn{1}{r}{--} & & \multicolumn{1}{r}{--} & \multicolumn{1}{r}{--} \\
& illegal & 1000 & -0.98 & & -0.95 & 0.00 & & -0.97 & -0.01 & & -0.93 & -0.01 \\
\hline
\end{tabular}
\caption{Editability on Othello split by whether the target board is legal, meaning some game under the variant's rules ends on exactly the flipped board with the same player to move, or illegal, meaning no game does. Both are decided exactly by an exhaustive search over games, and the cases it could not settle within its budget are left out (4 in \texttt{standard}, 51 in \texttt{adjacent-flip}). Each editor is read at its setting from \cref{tab:editability}. Without flipping, a token keeps its placer's color and players pass only when they cannot move, so each color's count is fixed by the number of moves and no board with a single flipped token is legal. The illegal rows of both no-flip variants are therefore the full bench of \cref{tab:editability}. Columns as in \cref{tab:editability}.}
\label{tab:legal_illegal}\end{table}

\begin{table}[ht]
\centering
\small
\setlength{\tabcolsep}{4pt}
\begin{tabular}{ll r E G EF G EF G EF}
\hline
\noalign{\vspace{0.25em}}
& & & \multicolumn{1}{c}{} & & \multicolumn{8}{c}{Editor} \\
\noalign{\vspace{0.25em}}
\cline{6-13}
\noalign{\vspace{0.25em}}
& & & \multicolumn{1}{c}{Unedited} & & \multicolumn{2}{c}{PI} & & \multicolumn{2}{c}{GS} & & \multicolumn{2}{c}{IM} \\
Variant & Target & \multicolumn{1}{c}{$n$} & \multicolumn{1}{c}{Index} & &
\multicolumn{1}{c}{Index} & \multicolumn{1}{c}{Fid.} & &
\multicolumn{1}{c}{Index} & \multicolumn{1}{c}{Fid.} & &
\multicolumn{1}{c}{Index} & \multicolumn{1}{c}{Fid.} \\
\hline
\noalign{\vspace{0.1em}}
\multirow{2}{*}{\texttt{standard}}
& legal   & 1000 & -0.92 & & +0.74 & 0.51 & & +0.78 & 0.73 & & +0.78 & 0.71 \\
& illegal & 1000 & -0.93 & & +0.72 & 0.61 & & +0.76 & 0.69 & & +0.76 & 0.69 \\
\cline{1-13}
\multirow{2}{*}{\texttt{adjacent-flip}}
& legal   & 1000 & -0.94 & & +0.31 & 0.03 & & +0.16 & 0.15 & & +0.71 & 0.56 \\
& illegal & 1000 & -0.95 & & +0.18 & 0.02 & & +0.05 & 0.08 & & +0.64 & 0.54 \\
\cline{1-13}
\multirow{2}{*}{\texttt{adjacent-noflip}}
& legal   & 1000 & -0.94 & & -0.04 & 0.18 & & -0.94 & -0.00 & & +0.88 & 0.78 \\
& illegal & 1000 & -0.95 & & -0.95 & 0.00 & & -0.95 & -0.00 & & +0.65 & 0.08 \\
\cline{1-13}
\multirow{2}{*}{\texttt{standard-noflip}}
& legal   & 0    & \multicolumn{1}{r}{--} & & \multicolumn{1}{r}{--} & \multicolumn{1}{r}{--} & & \multicolumn{1}{r}{--} & \multicolumn{1}{r}{--} & & \multicolumn{1}{r}{--} & \multicolumn{1}{r}{--} \\
& illegal & 1000 & -0.97 & & -0.97 & 0.00 & & -0.95 & -0.01 & & -0.86 & -0.08 \\
\hline
\end{tabular}
\caption{Editability on Othello for edits that flip two tokens, one black and one white, so that each color's count is unchanged. We generate 1000 reachable and unreachable cases for each variant, except for \texttt{standard-noflip}, as no such edit is legal there. Each editor is read at the setting the rule of \cref{tab:editability} picks within each group. Standard errors of the Edit Index are at most $0.02$. Columns as in \cref{tab:legal_illegal}.}
\label{tab:two_flip}\end{table}

\begin{revblock}
\paragraph{Fitting at a single time step.}
On \texttt{adjacent-noflip}, the inverse map fit across all time steps edits the reachable pairs of \cref{tab:two_flip} but not the single flips of \cref{tab:editability}. Fit at the edit step alone, it edits the single flips as well, even though no game could reach them, so unreachability is not what blocks these edits. The map is still brittle: it works at nearby steps where the same player moves, but fails at steps where the other player moves, and probes fit at the edit step still fail to edit (\cref{app:robustness}).

\paragraph{Writing where the token was placed.}
Writing a single-flip edit at the position where the flipped token was placed does not edit on its own. Writing it there and at the current step raises the inverse map's Edit Index to $+0.40$, still below \texttt{adjacent-flip} with a write at the current step alone. PI and GS fail in every configuration.
\end{revblock}

\FloatBarrier
\subsection{Matching GPT-style Prediction in Rayworld}
\label{app:tokenized_rw}

Othello and Rayworld models differ in their interface as well as their environment. The Othello model reads each move as a token and predicts a distribution over the next one, in the manner of a GPT-style language model. The Rayworld model reads and predicts frames as vectors under a squared error loss. To make sure this interface is not what separates the two environments, we train an \texttt{8-ray} Rayworld model with Othello's interface. With fixed reflectivities and no noise, every pixel takes one of three values depending on which disc it hits, or if it misses, so each frame is one of a finite set of patterns. Every pattern that occurs in the dataset becomes a token, giving a vocabulary of 421 frames plus a reserved unknown token. This allows us to match the setup exactly to the Othello model, with a token embedding in, a softmax over next frames out, and a cross-entropy loss, trained with the same recipe but on the \texttt{8-ray} frame model's sequences. Probes and editors act on the residual stream and carry over unchanged. Only scoring needs a choice, since the model outputs a distribution over frames rather than a frame. The frame-set reading scores the model's distribution over frames the way Othello scores its distribution over moves, with each world's true next frame in place of the set of legal moves. An edit scores $+1$ when all the probability moves onto the edited world's frame and $-1$ when it stays on the unedited one. The mean-frame reading averages the distribution into one frame and scores it exactly as the regular Rayworld models are scored. 

The interface changes almost nothing. The token model decodes the state as well as the frame model, with every trained probe within $0.013$ of it (\cref{tab:tokens_decodability}), and its mean frame predicts as accurately (\cref{tab:predictive_skill}). Its editability also follows the same pattern under either reading (\cref{tab:tokens_editability}). On continuous positions PI and GS fail while IM \rev{succeeds}, and on the categorical target the probe-derived editors improve while IM \rev{succeeds}. The frame-set reading gives lower scores to the editors that \rev{succeed}, since it credits only probability on the exact edited frame, and a prediction that places the disc one ray off earns nothing. Whatever keeps the probes from steering Rayworld's continuous positions, it is not the vector interface.

\begin{table}[ht]
\centering
\small
\setlength{\tabcolsep}{4pt}
\begin{tabular}{ll PP G PP G PP G PP}
\hline
\noalign{\vspace{0.25em}}
& & \multicolumn{11}{c}{Probe Type} \\
\noalign{\vspace{0.25em}}
\cline{3-13}
\noalign{\vspace{0.25em}}
& & \multicolumn{2}{c}{Observation} & & \multicolumn{2}{c}{Random init} & & \multicolumn{2}{c}{Trained} & & \multicolumn{2}{c}{Inverse Map MLP} \\
State & Model
& \multicolumn{1}{c}{Lin.} & \multicolumn{1}{c}{MLP} & &
\multicolumn{1}{c}{Lin.} & \multicolumn{1}{c}{MLP} & &
\multicolumn{1}{c}{Lin.} & \multicolumn{1}{c}{MLP} & &
\multicolumn{1}{c}{Best Point} & \multicolumn{1}{c}{Edit Point} \\
\hline
\noalign{\vspace{0.1em}}
\multirow{2}{*}{Continuous}
& \texttt{8-ray} frames & 0.329 & 0.877 & & 0.883 & 0.919 & & 0.886 & 0.932 & & 0.585 & 0.580 \\
& \texttt{8-ray} tokens & 0.329 & 0.877 & & 0.876 & 0.885 & & 0.899 & 0.935 & & 0.611 & 0.611 \\
\cline{1-13}
\multirow{2}{*}{Categorical}
& \texttt{8-ray} frames & 0.477 & 0.935 & & 0.922 & 0.935 & & 0.935 & 0.943 & & 0.618 & 0.615 \\
& \texttt{8-ray} tokens & 0.477 & 0.935 & & 0.921 & 0.925 & & 0.940 & 0.944 & & 0.641 & 0.641 \\
\hline
\end{tabular}
\caption{Decodability of the \texttt{8-ray} frame model and the \texttt{8-ray} token model, which sees the same sequences as a vocabulary of whole-frame tokens and predicts a distribution over next frames. Columns as in \cref{tab:decodability}. The observation baseline is shared, since both models see the same frames.}
\label{tab:tokens_decodability}
\end{table}

\begin{table}[ht]
\centering
\small
\setlength{\tabcolsep}{4pt}
\begin{tabular}{ll E G EF G EF G EF}
\hline
\noalign{\vspace{0.25em}}
& & \multicolumn{1}{c}{} & & \multicolumn{8}{c}{Editor} \\
\noalign{\vspace{0.25em}}
\cline{5-12}
\noalign{\vspace{0.25em}}
& & \multicolumn{1}{c}{Unedited} & & \multicolumn{2}{c}{PI} & & \multicolumn{2}{c}{GS} & & \multicolumn{2}{c}{IM} \\
State & Model & \multicolumn{1}{c}{Index} & &
\multicolumn{1}{c}{Index} & \multicolumn{1}{c}{Fid.} & &
\multicolumn{1}{c}{Index} & \multicolumn{1}{c}{Fid.} & &
\multicolumn{1}{c}{Index} & \multicolumn{1}{c}{Fid.} \\
\hline
\noalign{\vspace{0.1em}}
\multirow{3}{*}{Continuous}
& \texttt{8-ray} frames                       & -0.90 & & +0.21 & 0.00 & & -0.07 & 0.12 & & +0.71 & 0.73 \\
& \texttt{8-ray} tokens, frame set$^*$  & -0.75 & & +0.00 & 0.25 & & -0.01 & 0.24 & & +0.65 & 0.70 \\
& \texttt{8-ray} tokens, mean frame           & -0.90 & & +0.14 & 0.19 & & -0.08 & 0.15 & & +0.84 & 0.75 \\
\cline{1-12}
\multirow{3}{*}{Categorical}
& \texttt{8-ray} frames                       & -0.90 & & +0.38 & 0.05 & & +0.46 & 0.46 & & +0.87 & 0.73 \\
& \texttt{8-ray} tokens, frame set$^*$  & -0.73 & & +0.23 & 0.44 & & +0.35 & 0.53 & & +0.78 & 0.80 \\
& \texttt{8-ray} tokens, mean frame           & -0.90 & & +0.46 & 0.32 & & +0.48 & 0.43 & & +0.92 & 0.75 \\
\hline
\end{tabular}
\caption{Editability of the \texttt{8-ray} frame model and the \texttt{8-ray} token model through the continuous and categorical state targets. Columns as in \cref{tab:editability}. Rows marked $*$ score the token model's next-frame distribution against the frame sets of the edited and unedited worlds, the legal-set construction used for Othello. Mean-frame rows take the expectation of that distribution and score it as an \texttt{8-ray} frame model with the ray-zone construction, choosing each editor's setting by the rule of \cref{tab:editability} on that readout.}
\label{tab:tokens_editability}\end{table}

\FloatBarrier
\subsection{Additional Rayworld Variants}
\label{app:additional_rayworld_runs}

Two further variants investigate whether the probes fail on continuous positions because the observation shows position too crudely. In \texttt{standard} Rayworld a disc lights every ray it covers at its full reflectivity, so a frame changes only when a disc's edge crosses a ray, and a model might predict well without ever tracking a precise position. The \texttt{smooth} variant shades each disc like a dome. A ray reads the disc's full reflectivity where it passes through the center and fades smoothly to zero toward the edge, following $(1 - (d/r)^2)^2$ for a ray at distance $d$ from the center of a disc of radius $r$. Any small shift of the disc then changes the brightness of every ray it lights, not just the rays at its edges, while the set of rays it covers stays the same. The \texttt{obs5} variant keeps coarse rays but adds multiple viewpoints. Five \texttt{8-ray} observers stand evenly spaced on a ring around a circular arena, and each frame joins their five views into 40 values. A disc hidden from one observer behind the other disc stays visible to the rest, so together the views pin down its position far better than any one of them. Each variant trains the same model with the same recipe on data of its own, and we probe the same continuous Cartesian state.

Both variants make position easier to read, and neither makes it easier to steer (\cref{tab:additional_rw}). On \texttt{smooth}, training lifts the linear probe from $0.62$ on a randomly initialized model to $0.96$, the largest gap of any Rayworld variant, so the model does learn a linearly readable position. On \texttt{obs5} an MLP reads position from the raw frames alone at $0.98$, and both probes on the trained model exceed $0.97$. Yet PI and GS fail on both variants, with Edit Indices between $-0.19$ and $+0.02$ at an Edit Fidelity of $0.02$, while IM \rev{succeeds} on both at least as well as on \texttt{standard}. Therefore, Rayworld's outcomes are not due to a frame that reveals position too coarsely, since the smooth profile makes every lit ray track the exact position, nor are they due to a single viewpoint that leaves position ambiguous, since five viewpoints pin it down.

A third variant, \texttt{pair}, adds an invariant to the world. Its two discs move as a rigid pair, at a fixed distance of $2.0$, with a shared velocity. However, the edit bench still teleports a single disc as usual, breaking the rigid pair. We expected these edits to fail, since the model never sees a disc move apart from its partner. In fact, the world state is now redundant, as the positions can be represented equally well with three dimensions (one 2D coordinate, plus an angle). Yet IM still \rev{succeeds} from the Cartesian input, at $+0.41$ with an Edit Fidelity of $0.30$ against $+0.59$ and $0.66$ on \texttt{standard}, which suggests that the environment's invariants influence editability but do not dictate it.

\begin{table}[ht]
\centering
\small
\setlength{\tabcolsep}{4pt}

(a) Probe Skill\\[0.35em]
\begin{tabular}{l PP G PP G PP G PP}
\hline
\noalign{\vspace{0.25em}}
& \multicolumn{11}{c}{Probe Type} \\
\noalign{\vspace{0.25em}}
\cline{2-12}
\noalign{\vspace{0.25em}}
& \multicolumn{2}{c}{Observation} & & \multicolumn{2}{c}{Random init} & & \multicolumn{2}{c}{Trained} & & \multicolumn{2}{c}{Inverse Map MLP} \\
Variant
& \multicolumn{1}{c}{Lin.} & \multicolumn{1}{c}{MLP} & &
\multicolumn{1}{c}{Lin.} & \multicolumn{1}{c}{MLP} & &
\multicolumn{1}{c}{Lin.} & \multicolumn{1}{c}{MLP} & &
\multicolumn{1}{c}{Best Point} & \multicolumn{1}{c}{Edit Point} \\
\hline
\noalign{\vspace{0.1em}}
\texttt{standard} & 0.305 & 0.885 & & 0.738 & 0.945 & & 0.872 & 0.973 & & 0.741 & 0.339 \\
\texttt{smooth}   & 0.152 & 0.829 & & 0.615 & 0.949 & & 0.962 & 0.994 & & 0.885 & 0.746 \\
\texttt{obs5}     & 0.575 & 0.981 & & 0.934 & 0.990 & & 0.979 & 0.997 & & 0.641 & 0.488 \\
\texttt{pair}     & 0.502 & 0.890 & & 0.800 & 0.969 & & 0.922 & 0.994 & & 0.765 & 0.373 \\
\hline
\end{tabular}

\vspace{1em}

(b) Edit Index and Edit Fidelity\\[0.35em]
\begin{tabular}{l E G EF G EF G EF}
\hline
\noalign{\vspace{0.25em}}
& \multicolumn{1}{c}{} & & \multicolumn{8}{c}{Editor} \\
\noalign{\vspace{0.25em}}
\cline{4-11}
\noalign{\vspace{0.25em}}
& \multicolumn{1}{c}{Unedited} & & \multicolumn{2}{c}{PI} & & \multicolumn{2}{c}{GS} & & \multicolumn{2}{c}{IM} \\
Variant & \multicolumn{1}{c}{Index} & &
\multicolumn{1}{c}{Index} & \multicolumn{1}{c}{Fid.} & &
\multicolumn{1}{c}{Index} & \multicolumn{1}{c}{Fid.} & &
\multicolumn{1}{c}{Index} & \multicolumn{1}{c}{Fid.} \\
\hline
\noalign{\vspace{0.1em}}
\texttt{standard} & -0.93 & & -0.10 & 0.04 & & -0.16 & 0.04 & & +0.59 & 0.66 \\
\texttt{smooth}   & -0.97 & & -0.09 & 0.02 & & -0.19 & 0.02 & & +0.67 & 0.74 \\
\texttt{obs5}     & -0.89 & & +0.02 & 0.02 & & -0.10 & 0.02 & & +0.60 & 0.66 \\
\texttt{pair}     & -0.95 & & -0.10 & 0.01 & & -0.15 & 0.01 & & +0.41 & 0.30 \\
\hline
\end{tabular}
\caption{Decodability and editability of the three additional Rayworld variants, \texttt{smooth} (the rendering profile falls off continuously from each disc's center), \texttt{obs5} (five \texttt{8-ray} observers spaced around the arena) and \texttt{pair} (the two discs move as a rigid pair), beside \texttt{standard} Rayworld for reference. All four use the continuous position target. Panel (a) reports Probe Skill and panel (b) reports Edit Index and Edit Fidelity.}
\label{tab:additional_rw}
\end{table}

\FloatBarrier
\subsection{Inverse Mapping Trends Across Residual Points}
\label{app:im_editability_trends}

In Othello the inverse map successfully edits at only two residual points, and the way its fit changes with depth suggests why. \Cref{tab:im_by_point} tracks editability of the \texttt{standard} Othello and \texttt{standard} Rayworld models through all nine residual points. In Othello, the first five blocks look like a compression phase. The inverse map's $R^2$ climbs from $0.39$ at the embedding to $0.83$ at point 5, as the board comes to determine more and more of the latent state, then dips to $0.73$ at point 7. IM \rev{succeeds} only at the top of this climb, at points 4 and 5, and at every other point its write leaves the prediction worse than no edit. A reconstruction test further supports this. Overwriting the latent state with the inverse mapping of the \textit{unedited} board state, which should change nothing, keeps the output's error low only at points 4 and 5, and raises it substantially elsewhere. We read this as a bottleneck. The board is a sufficient statistic of the history for the next move, since the legal moves depend on the history only through it, and with flipping it is a complicated function of that history. The model appears to assemble the board in its early blocks, hold it near points 4 and 5, and then expand it into the features of its prediction. \rev{Restricting attention further supports this. If each position attends only to itself from block 7 onward, so that those blocks cannot read the history, the model's loss barely changes ($+0.013$ nats). From block 6 onward this costs a little ($+0.076$ nats), and from block 5 or earlier it costs clearly more ($> +0.258$ nats). By about point 5 the model has gathered what it needs from the history, and its last blocks compute the prediction from the current position alone.} A board written earlier meets blocks that are still gathering one from the moves, and a board written later lacks the detail those blocks have added since. The inverse $R^2$ alone does not locate the bottleneck, since it is higher at point 8 than at point 5 while IM fails there.

Rayworld shows no such bottleneck. The inverse map's $R^2$ is highest at the embedding, where the latent state is a direct function of the current frame, and falls to about $0.3$ from point 1 on, where it stays. Yet IM improves the prediction at every point, and its Edit Index climbs steadily to $+0.59$ at point 6, at an Edit Fidelity of $0.66$, and stays above $+0.54$ through point 8. No single depth holds a compact state that the rest of the network depends on, and the write works at most depths. \rev{Instead, the model appears to refine a rendering of the next frame across its layers (\cref{app:inverse_controls}).} The contrast fits what each environment demands of prediction \rev{(\cref{sec:forcing_state})}. Othello's rules make the board both necessary \rev{for prediction and complexly related to the moves}, which gives a model reason to form it once and route everything through it. Rayworld's frames already show most of the state, and its dynamics are linear, so its model faces no such pressure. The probe-derived editors follow a similar split. On Othello, PI and GS peak at point 4, just before IM, and fail from point 6 on, while on Rayworld they fail at every point.

\begin{table}[ht]
\centering
\small
\setlength{\tabcolsep}{4pt}
\begin{tabular}{l PP EF G PP EF}
\hline
\noalign{\vspace{0.25em}}
& \multicolumn{4}{c}{Othello \texttt{standard}} & & \multicolumn{4}{c}{Rayworld \texttt{standard}} \\
\noalign{\vspace{0.25em}}
\cline{2-5} \cline{7-10}
\noalign{\vspace{0.25em}}
& \multicolumn{1}{c}{Probe} & \multicolumn{1}{c}{Inverse} & \multicolumn{2}{c}{IM} & & \multicolumn{1}{c}{Probe} & \multicolumn{1}{c}{Inverse} & \multicolumn{2}{c}{IM} \\
Residual point & \multicolumn{1}{c}{Skill} & \multicolumn{1}{c}{$R^2$} & \multicolumn{1}{c}{Index} & \multicolumn{1}{c}{Fid.} & &
\multicolumn{1}{c}{Skill} & \multicolumn{1}{c}{$R^2$} & \multicolumn{1}{c}{Index} & \multicolumn{1}{c}{Fid.} \\
\hline
\noalign{\vspace{0.1em}}
0 (embedding) & 0.346 & 0.393 & -0.23 & -1.07 & & 0.940 & 0.741 & -0.23 & 0.25 \\
1             & 0.810 & 0.682 & -0.20 & -0.99 & & 0.965 & 0.340 & +0.06 & 0.43 \\
2             & 0.879 & 0.769 & -0.03 & -0.64 & & 0.968 & 0.310 & +0.29 & 0.54 \\
3             & 0.927 & 0.796 & +0.26 & -0.14 & & 0.969 & 0.284 & +0.44 & 0.61 \\
4             & 0.952 & 0.818 & +0.57 & 0.51 & & 0.970 & 0.298 & +0.54 & 0.67 \\
5             & 0.966 & 0.827 & +0.81 & 0.62 & & 0.971 & 0.343 & +0.57 & 0.66 \\
6             & 0.974 & 0.753 & +0.39 & -0.21 & & 0.972 & 0.339 & +0.59 & 0.66 \\
7             & 0.975 & 0.728 & +0.22 & -1.03 & & 0.972 & 0.342 & +0.58 & 0.64 \\
8             & 0.976 & 0.884 & +0.16 & -2.08 & & 0.973 & 0.344 & +0.55 & 0.60 \\
\hline
\end{tabular}
\caption{Decodability and inverse-map editability at every residual point of the \texttt{standard} Othello and \texttt{standard} Rayworld models. Probe Skill is the MLP probe's, the inverse $R^2$ is the held-out fit of $g$ from simulator state to latent state at that point, and the IM columns give the Edit Index and Edit Fidelity of the inverse map overwrite $z^* = g(s^*)$ written at that point. Point $k$ is the residual stream after block $k$.}
\label{tab:im_by_point}\end{table}

\FloatBarrier
\subsection{Appearance Bins Against Evenly Spaced Bins}
\label{app:rw_grid_ablation}

For the probe-derived editors, categorical targets seem to help even when their bins ignore the observation, and they help most when the bins are matched to it. We compare the appearance target on the \texttt{8-ray} model against evenly spaced grids over the frustum coordinates, matched to it at 30 bins and increased beyond it at 128. Every bin is labeled empty, bright disc, or dim disc (\cref{tab:categorical}). Every one of these grids improves GS from the continuous position target's Edit Index of $-0.07$ at an Edit Fidelity of $0.12$, reaching $+0.35$ to $+0.39$ at Edit Fidelities of at least $0.33$. Labeled the same way, the appearance bins improve it further, to $+0.59$ at an Edit Fidelity of $0.56$, and their factorized form from the main tables also beats every uniform grid. Much of the gain is therefore in binning at all, and the rest in choosing bins the observation already defines. PI trends the same way, albeit less cleanly. None of this tracks Probe Skill, since the grids decode worse than the position target and still edit better than it. 

Snapping each position to the center of its appearance bin while keeping the regression read-out returns GS to $-0.05$ at an Edit Fidelity of $0.07$, the continuous result, even though the partition there is the one that works for the categorical state. The inverse map tends to edit well in every Rayworld categorical grid variant, but similarly does best when the grid matches the observation space. 

\begin{table}[ht]
\centering
\small
\setlength{\tabcolsep}{4pt}
\begin{tabular}{l PP G EF G EF G EF}
\hline
\noalign{\vspace{0.25em}}
& \multicolumn{2}{c}{Probe Skill} & & \multicolumn{8}{c}{Editor} \\
\noalign{\vspace{0.25em}}
\cline{2-3} \cline{5-12}
\noalign{\vspace{0.25em}}
& \multicolumn{1}{c}{} & \multicolumn{1}{c}{} & \multicolumn{1}{c}{} & \multicolumn{2}{c}{PI} & & \multicolumn{2}{c}{GS} & & \multicolumn{2}{c}{IM} \\
Probe target (\texttt{8-ray} model) & \multicolumn{1}{c}{Lin.} & \multicolumn{1}{c}{MLP} & &
\multicolumn{1}{c}{Index} & \multicolumn{1}{c}{Fid.} & &
\multicolumn{1}{c}{Index} & \multicolumn{1}{c}{Fid.} & &
\multicolumn{1}{c}{Index} & \multicolumn{1}{c}{Fid.} \\
\hline
\noalign{\vspace{0.1em}}
Position, regression     & 0.886 & 0.932 & & +0.21 & 0.00 & & -0.07 & 0.12 & & +0.71 & 0.73 \\
Appearance (30 bins)  & 0.893 & 0.904 & & +0.44 & 0.11 & & +0.59 & 0.56 & & +0.88 & 0.73 \\
Appearance, factorized   & 0.935 & 0.943 & & +0.38 & 0.05 & & +0.46 & 0.46 & & +0.87 & 0.73 \\
Grid $6 \times 5$ (30)   & 0.466 & 0.535 & & +0.30 & 0.02 & & +0.35 & 0.33 & & +0.64 & 0.54 \\
Grid $10 \times 3$ (30)  & 0.443 & 0.496 & & +0.37 & 0.09 & & +0.35 & 0.35 & & +0.70 & 0.57 \\
Grid $16 \times 8$ (128) & 0.156 & 0.212 & & -0.05 & 0.06 & & +0.39 & 0.34 & & +0.73 & 0.64 \\
Snapped regression (30)  & 0.959 & 0.991 & & +0.29 & 0.00 & & -0.05 & 0.07 & & +0.70 & 0.69 \\
\hline
\end{tabular}
\caption{Probe targets on the Rayworld \texttt{8-ray} model, each a new probe fit on the same trained model with the same editors run through it on 1000 cases. Grids are evenly spaced bins over the frustum coordinates, as size-matched and resolution-matched controls for the appearance bins. Snapped regression replaces each position by the center of its appearance bin and keeps the regression pipeline. Columns as in \cref{tab:decodability,tab:editability}.}
\label{tab:categorical}\end{table}

\FloatBarrier
\subsection{Comparing Inverse Mapping to Nearest Neighbor Replacement}
\label{app:inverse_nearest_neighbor}

An inverse map fit on about 1.2 million pairs could succeed by memorizing them, recalling the latent state of a training example whose simulator state lies close to the target. We test this against a nearest-neighbor lookup with nothing to fit. Given a target simulator state, it finds the ten training examples whose simulator states are closest to it, takes the model's latent states on those examples, and writes their average into the model in place of the current latent state. Closeness is the number of differing cells between Othello boards, and the standardized Euclidean distance between Rayworld states. It draws on the same fit set as the inverse map, is swept over the same residual points, and is scored the same way. To compare how well each map fits, we also predict the latent states of held-out examples from their simulator states and report each map's latent $R^2$ at the point where the inverse map edits.

In every variant the inverse map beats the nearest neighbor at predicting latent states. It is also a better editor on every single Rayworld variant, while in Othello the nearest neighbor replacement never successfully edits. The inverse map therefore does more than recall nearby examples.

\begin{table}[ht]
\centering
\small
\setlength{\tabcolsep}{4pt}
\begin{tabular}{ll PP G EF G EF}
\hline
\noalign{\vspace{0.25em}}
& & \multicolumn{2}{c}{Latent $R^2$} & & \multicolumn{5}{c}{Editor} \\
\noalign{\vspace{0.25em}}
\cline{3-4} \cline{6-10}
\noalign{\vspace{0.25em}}
& & \multicolumn{1}{c}{} & \multicolumn{1}{c}{} & \multicolumn{1}{c}{} & \multicolumn{2}{c}{Inverse Map} & & \multicolumn{2}{c}{Nearest Neighbor} \\
Class & Variant & \multicolumn{1}{c}{IM} & \multicolumn{1}{c}{NN} & &
\multicolumn{1}{c}{Index} & \multicolumn{1}{c}{Fid.} & &
\multicolumn{1}{c}{Index} & \multicolumn{1}{c}{Fid.} \\
\hline
\noalign{\vspace{0.1em}}
\multirow{4}{*}{Othello}
& \texttt{standard}        & 0.827 & 0.726 & & +0.81 & 0.62 & & -0.09 & -0.81 \\
& \texttt{adjacent-flip}   & 0.959 & 0.901 & & +0.66 & 0.49 & & -0.70 & -0.27 \\
& \texttt{adjacent-noflip} & 0.855 & 0.750 & & -0.03 & 0.26 & & -0.82 & -0.05 \\
& \texttt{standard-noflip} & 0.855 & 0.762 & & -0.93 & -0.01 & & -0.90 & -0.07 \\
\cline{1-10}
\multirow{6}{*}{Rayworld}
& \texttt{standard} & 0.339 & 0.068 & & +0.59 & 0.66 & & +0.33 & 0.24 \\
& \texttt{blink}    & 0.338 & 0.059 & & +0.51 & 0.55 & & +0.27 & 0.22 \\
& \texttt{128-ray} (big discs) & 0.321 & 0.143 & & +0.57 & 0.68 & & +0.38 & 0.37 \\
& \texttt{16-ray}   & 0.516 & 0.357 & & +0.66 & 0.71 & & +0.49 & 0.45 \\
& \texttt{8-ray}    & 0.580 & 0.424 & & +0.71 & 0.73 & & +0.54 & 0.47 \\
& \texttt{5-ray}    & 0.469 & 0.263 & & +0.81 & 0.77 & & +0.60 & 0.54 \\
\hline
\end{tabular}
\caption{The inverse map $g$ against the nearest-neighbor control, which overwrites the latent state with the mean latent state of the nearest training states to the target. Latent $R^2$ is each map's held-out fit of the latent state at the residual point where the inverse map edits.}
\label{tab:im_vs_nn}\end{table}

\FloatBarrier
\subsection{Persistent Transformer Edits with History Rewriting}
\label{app:history_editing}

Transformers make persistent edits hard, because they carry no state from one step to the next. An edit written at the last position changes the next prediction, but at the following step the model reads the original history again, and all that remains of the edit is the one predicted frame now appended to it. That frame shows the disc in its new place while every frame before it shows the disc on its old path, and the model sides with the history. On the \texttt{standard} Rayworld model, a single inverse-map write \rev{succeeds} at the edit frame, at $+0.61$ with an Edit Fidelity of $0.64$ on the 32 cases we use here, yet one step later the rollout has largely returned to the unedited world, at $-0.63$. Scoring rollouts would therefore mix two questions: whether the model reads the written state and whether it carries that state forward against a contradicting history. The main paper asks only the first, and scores every edit on the next prediction.

We introduce history rewriting to make the history agree with the edit. The edited state keeps the moved disc's velocity, and Rayworld's dynamics are reversible, so integrating backward from the target gives the counterfactual state at every earlier frame, with the other disc on its original path. We then rebuild the history one frame at a time. At each step the model reads the frames rewritten so far, we overwrite its latent state at the last position with $g$ of the counterfactual state, and the model's prediction becomes the next rewritten frame. The first frame is kept, since nothing precedes it. The rollout then starts from the rewritten history. The simulator never renders an observation in this process. Every rewritten frame comes from the model, driven by written states alone. 

\begin{figure}[ht]
\centering
\includegraphics[width=1.0\linewidth]{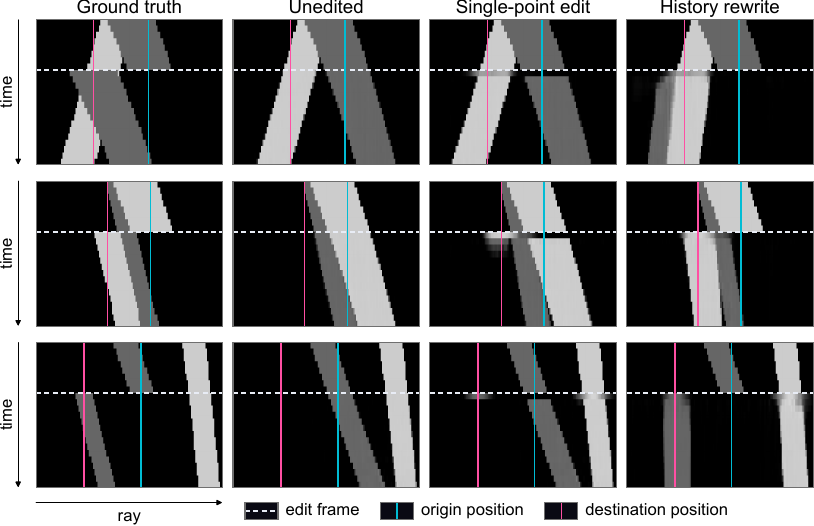}
\caption{Persistent edits on \texttt{standard} Rayworld through history rewriting, one edit per row with time running downward. Every column shows the same last eight observed frames above the dashed line. Below it are the edited world's true continuation and 15-step rollouts from the unedited model, from a single inverse-map write at the edit frame, and from the rewritten history with the same write. Cyan and magenta lines mark the edited disc's origin and destination at the edit frame.}
\label{fig:history_rewriting}
\end{figure}

The rewritten history successfully holds an edit through its rollout (\cref{fig:history_rewriting}). For a transformer, the history is part of the state. An edit persists only when the history agrees with it, and history rewriting achieves this with nothing but the inverse map and the model itself. It points toward driving a generative model continuously from a simulator, writing the simulator's state into the model at every step and using the model's own predictions as the observations, so that the simulator supplies the state and the model the detail. Two obstacles remain. Firstly, the backward integration relies on Rayworld's simple, reversible dynamics. Secondly, the rewritten frames also drift from the true render, so errors may compound over longer horizons. A model that carries its state forward explicitly, as a recurrent model does, might avoid rewriting the history at all.

\FloatBarrier
\subsection{Selecting for Edit Fidelity Instead of Edit Index}
\label{app:edit_fidelity_choice}

Our main tables choose each editor's setting by its Edit Index, among the settings that leave the prediction at least as accurate as no edit. The rule rewards the strongest edit that does no harm, but it could obscure an editor sitting right at that cutoff, or make our conclusions depend on where the cutoff lies. \Cref{tab:fidelity_selected} applies an alternative rule, choosing the setting with the highest Edit Fidelity outright and reporting the Edit Index reached there. This rule prioritizes accuracy over change. PI and GS mostly move to smaller steps, trading Edit Index for Edit Fidelity, while IM, which has no step size, moves by at most $0.10$ in Edit Index.

Every conclusion of the main text survives the change of rule. All three editors still edit \texttt{standard} Othello. On \texttt{adjacent-flip} only IM \rev{succeeds}, and on the no-flip variants no editor does. In Rayworld, PI and GS fail on continuous positions in every variant, while IM \rev{succeeds} in all of them and rises as the rays coarsen. On the categorical target both probe-derived editors again rise as the rays coarsen, and IM \rev{succeeds} throughout.

\begin{table}[ht]
\centering
\small
\setlength{\tabcolsep}{4pt}
\begin{tabular}{ll E G EF G EF G EF}
\hline
\noalign{\vspace{0.25em}}
& & \multicolumn{1}{c}{} & & \multicolumn{8}{c}{Editor} \\
\noalign{\vspace{0.25em}}
\cline{5-12}
\noalign{\vspace{0.25em}}
& & \multicolumn{1}{c}{Unedited} & & \multicolumn{2}{c}{PI} & & \multicolumn{2}{c}{GS} & & \multicolumn{2}{c}{IM} \\
Class & Variant & \multicolumn{1}{c}{Index} & &
\multicolumn{1}{c}{Index} & \multicolumn{1}{c}{Fid.} & &
\multicolumn{1}{c}{Index} & \multicolumn{1}{c}{Fid.} & &
\multicolumn{1}{c}{Index} & \multicolumn{1}{c}{Fid.} \\
\hline
\noalign{\vspace{0.1em}}
\multirow{4}{*}{Othello}
& \texttt{standard}        & -0.93 & & +0.82 & 0.70 & & +0.78 & 0.73 & & +0.81 & 0.62 \\
& \texttt{adjacent-flip}   & -0.96 & & +0.03 & 0.36 & & -0.06 & 0.21 & & +0.56 & 0.61 \\
& \texttt{adjacent-noflip} & -0.96 & & -0.23 & 0.19 & & -0.96 & -0.01 & & -0.03 & 0.26 \\
& \texttt{standard-noflip} & -0.98 & & -0.95 & 0.00 & & -0.97 & -0.01 & & -0.93 & -0.01 \\
\cline{1-12}
\multirow{6}{*}{\shortstack[l]{Rayworld\\(continuous state)}}
& \texttt{standard} & -0.93 & & -0.45 & 0.13 & & -0.59 & 0.11 & & +0.54 & 0.67 \\
& \texttt{blink}    & -0.92 & & -0.36 & 0.14 & & -0.49 & 0.13 & & +0.49 & 0.55 \\
& \texttt{128-ray} (big discs) & -0.94 & & -0.47 & 0.16 & & -0.29 & 0.16 & & +0.55 & 0.68 \\
& \texttt{16-ray}   & -0.91 & & -0.15 & 0.19 & & -0.30 & 0.15 & & +0.66 & 0.71 \\
& \texttt{8-ray}    & -0.90 & & -0.16 & 0.22 & & -0.38 & 0.16 & & +0.71 & 0.74 \\
& \texttt{5-ray}    & -0.89 & & +0.03 & 0.26 & & -0.33 & 0.16 & & +0.81 & 0.77 \\
\cline{1-12}
\multirow{4}{*}{\shortstack[l]{Rayworld\\(categorical state)}}
& \texttt{128-ray} (3736 bins) & -0.94 & & -0.63 & 0.10 & & +0.10 & 0.38 & & +0.63 & 0.73 \\
& \texttt{16-ray} (77 bins)    & -0.91 & & -0.42 & 0.14 & & +0.12 & 0.36 & & +0.78 & 0.72 \\
& \texttt{8-ray} (30 bins)     & -0.90 & & +0.06 & 0.35 & & +0.40 & 0.46 & & +0.84 & 0.74 \\
& \texttt{5-ray} (14 bins)     & -0.91 & & +0.38 & 0.42 & & +0.58 & 0.55 & & +0.90 & 0.75 \\
\hline
\end{tabular}
\caption{\Cref{tab:editability} changed to the highest-fidelity selection rule. For each editor we report the residual point and step size with the highest Edit Fidelity, and give the Edit Index reached at that setting, instead of the highest Edit Index among settings whose Edit Fidelity is at least $0$. Rows, columns, and cases as in \cref{tab:editability}.}
\label{tab:fidelity_selected}\end{table}

\FloatBarrier
\subsection{Implementation Details}
\label{app:implementation}

\paragraph{Environments.}
The Rayworld frustum spans depths 3 to 12, and its width at each depth equals that depth. Each disc's initial depth is drawn, then its lateral position across the frustum at that depth, and it moves in a random direction at a speed between $0.05$ and $0.12$ per frame. The depth, lateral position, velocity direction, and speed are all drawn uniformly. A sequence is redrawn if either disc ever leaves the frustum or the two discs' centers ever come within $1.6$ diameters of each other, so every trajectory is a straight line at constant speed. Discs have radius $0.5$, or $1.0$ in the \texttt{N-ray} family, which casts $N + 2$ rays and drops the two that run along the walls. In \texttt{blink}, after the first three frames each disc begins a blackout with probability $0.05$ per frame, lasting a geometric number of frames with mean 7, capped at 12. Othello games come from the generator of \citet{li2023othello} under each variant's rules. A player passes only when they have no legal move, and a game ends when neither player can move. Each edit recolors one occupied square outside the four central ones, and a case is rejected if the flip leaves the legal moves unchanged or empty. Training, probe, evaluation, and edit sequences come from disjoint seed ranges.

\paragraph{Models and training.}
The transformer is the minGPT model of \citet{li2023othello}, with 8 heads, learned absolute position embeddings, and dropout of $0.1$. It reads 39 frames in Rayworld and 59 moves in Othello, whose vocabulary has 61 tokens. We train with AdamW at a learning rate of $10^{-3}$ and weight decay of $10^{-4}$, clip gradients at norm $1$, and hold the learning rate constant after 2,000 warmup steps. Main runs use seed 0 and the replicates use seeds 1 and 2. Each model trains on a single NVIDIA RTX 4090 or 5090, in 8 to 12 hours for Rayworld and 19 to 26 hours for Othello.

\paragraph{Probes and inverse map.}
Probes are fit at every position of each sequence, with an 80/20 split by sequence into fit and held-out sets. Linear regression probes are solved in closed form. The MLP probes, the linear categorical probes, and the inverse map train with Adam at a learning rate of $10^{-3}$ and batch size 4,096, for 200 epochs, or 50 epochs on the larger corpus of the categorical Rayworld targets. The inverse map minimizes squared error in the standardized latent space. Its input is the eight-dimensional state in Rayworld, the one-hot mine/theirs board in Othello, and for the categorical target the one-hot bin labels together with both discs' velocities. The nearest-neighbor control averages the latent states of the 10 fit states nearest the target, by standardized Euclidean distance on continuous states and Hamming distance on categorical ones. The random-init baseline uses the same budget as the trained model's probes. The observation baseline instead reads the whole input history, a much wider input that overfits this budget and so understates what the input reveals, so we fit it on 250,000 Rayworld sequences, 200,000 for the categorical targets, or 170,000 Othello games, for 50 epochs, where its gap between fit and held-out skill nearly vanishes. Across the runs in our tables the fit and held-out skills differ by at most $0.003$ for the linear probes and $0.005$ for the MLP probes, and the inverse map's two $R^2$ values agree to three decimals. 

\paragraph{Editors and evaluation.}
Every editor writes only at the last position of the history. On continuous targets PI sweeps $\alpha \in \set{0.1, 0.25, 0.5, 0.75, 1, 1.5, 2, 3, 5, 8, 12, 20, 35, 60, 100, 175}$ and GS sweeps $\eta \in \set{0.002, 0.005, 0.01, 0.02, 0.05, 0.1, 0.2, 0.35, 0.5, 0.7}$. On categorical targets, Othello's included, PI sweeps $\alpha \in \set{0.25, 0.5, 1, 1.5, 2, 3, 5, 10, 20, 35, 60, 100}$ and GS sweeps $\eta \in \set{0.05, 0.1, 0.2, 0.35, 0.7, 1.5}$. GS starts at residual point 0, 2, 4, 6, or 8, while PI and IM are swept over all nine points. Edit Index is averaged over cases. Edit Fidelity divides errors that are each averaged over all cases, rather than averaging a ratio per case.

\FloatBarrier
\subsection{Additional Qualitative Editability Visualizations}
\label{app:additional_edit_visuals}

The following figures show more edits in the style of \cref{fig:qualitative_edits}. In each Othello figure, every column is one randomly drawn edit for one rule variant. The top two boards show the legal moves before and after the flip, and the rest show the model's prediction after each editor's write, with shading that saturates at 2\%. Outlines mark the flipped token and the cells whose legality it changes. Where no setting reaches an Edit Fidelity of $0$ the figures show the setting with the highest Edit Index instead of the table's fallback. Each Rayworld panel shows one world under every variant, with single next-step predictions rather than rollouts, each above a strip of its error against the edited ground truth. We drew only teleports that visibly change the \texttt{5-ray} frame. The categorical IM is blank for \texttt{standard} and \texttt{blink}, which have no categorical inverse map.

\begin{figure}[ht]
\centering
\includegraphics[width=0.9\linewidth]{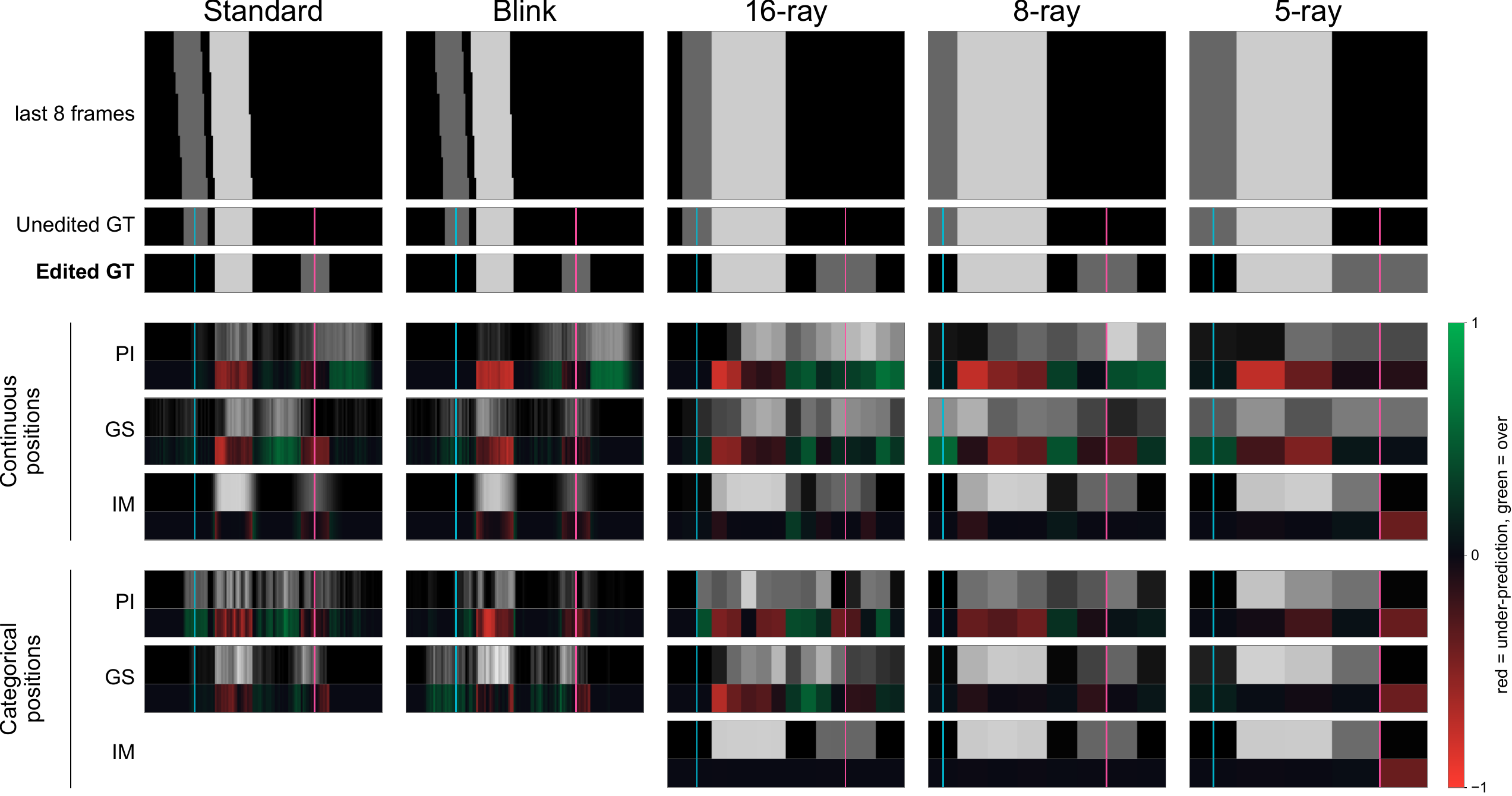}

\vspace{1em}

\includegraphics[width=0.9\linewidth]{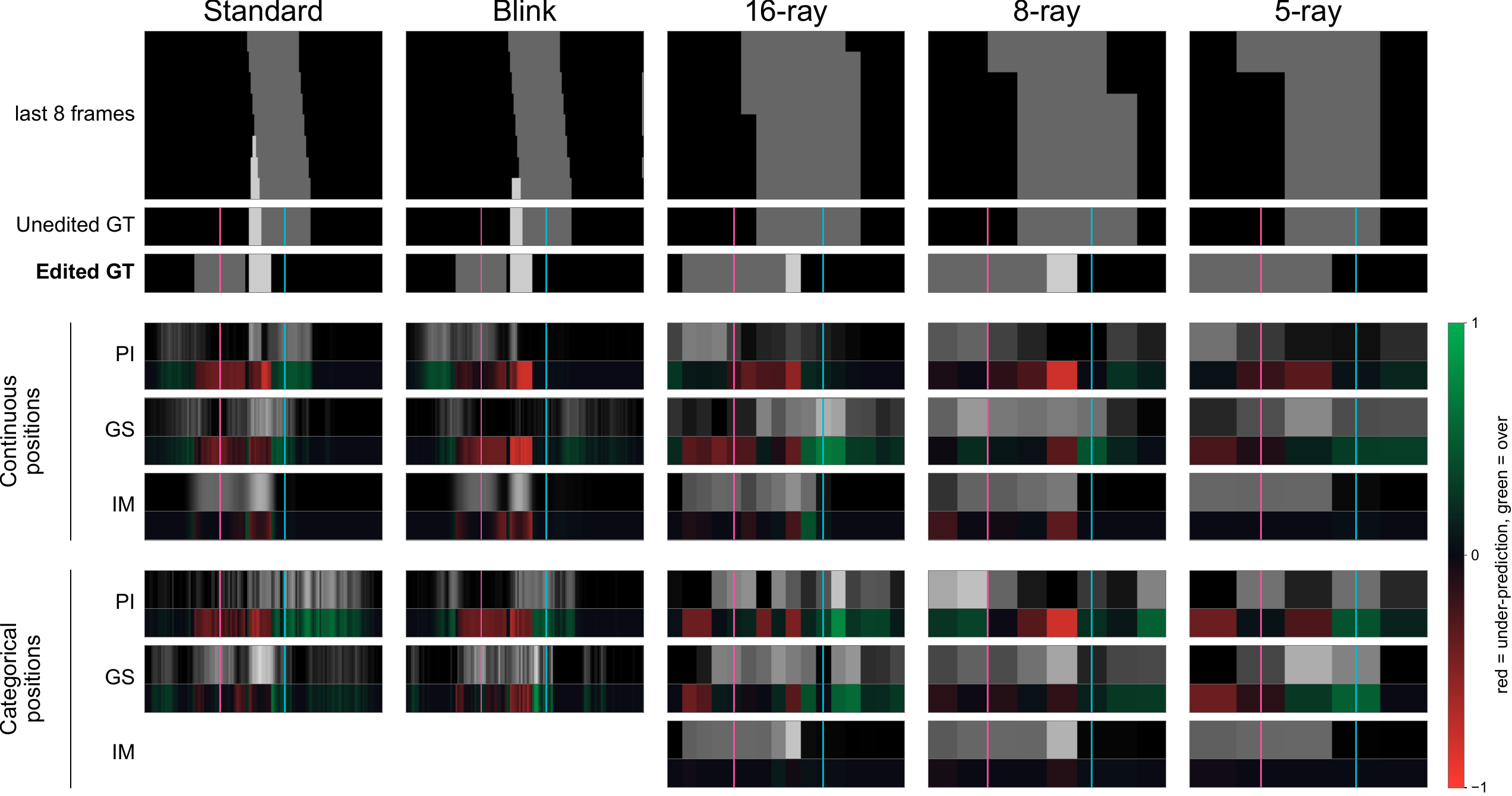}

\vspace{1em}

\includegraphics[width=0.9\linewidth]{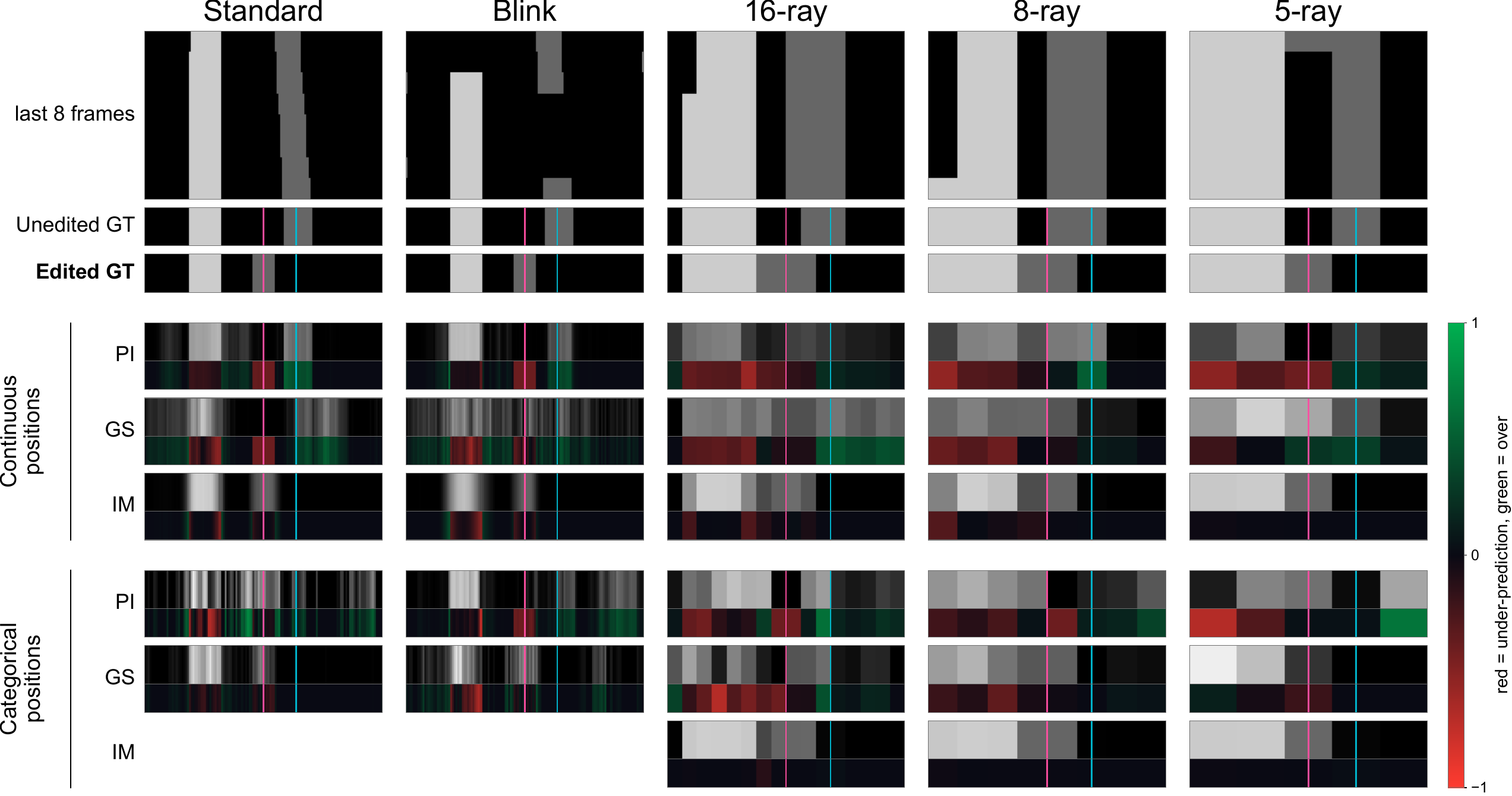}
\caption{Additional qualitative visualizations of Rayworld edits.}
\label{fig:more_qualitative_edits_rayworld_1}
\end{figure}
\begin{figure}[ht]
\centering
\includegraphics[width=1.0\linewidth]{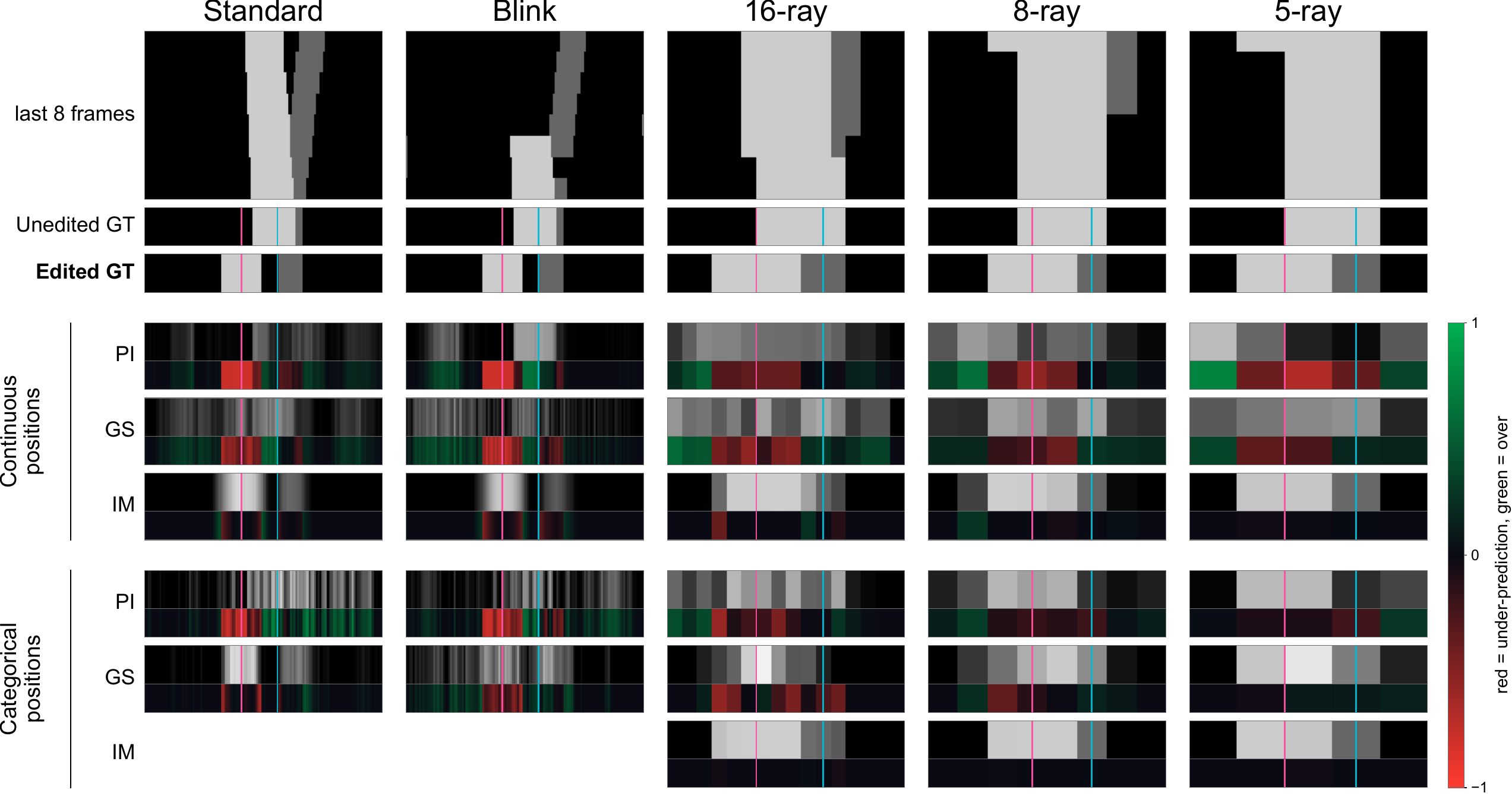}

\vspace{1em}

\includegraphics[width=1.0\linewidth]{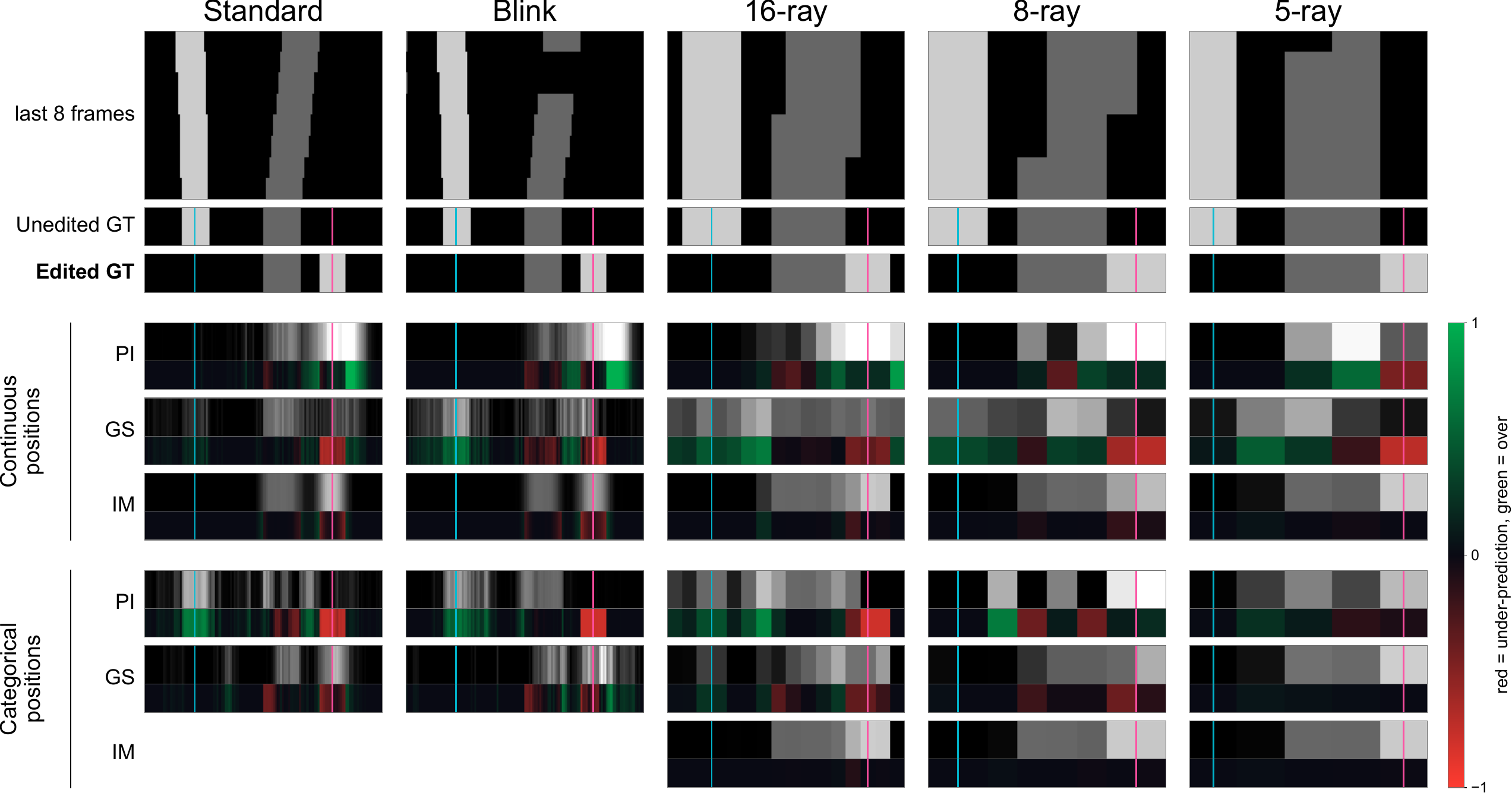}
\caption{Additional qualitative visualizations of Rayworld edits.}
\label{fig:more_qualitative_edits_rayworld_2}
\end{figure}

\begin{figure}[ht]
\centering
\includegraphics[width=1.0\linewidth]{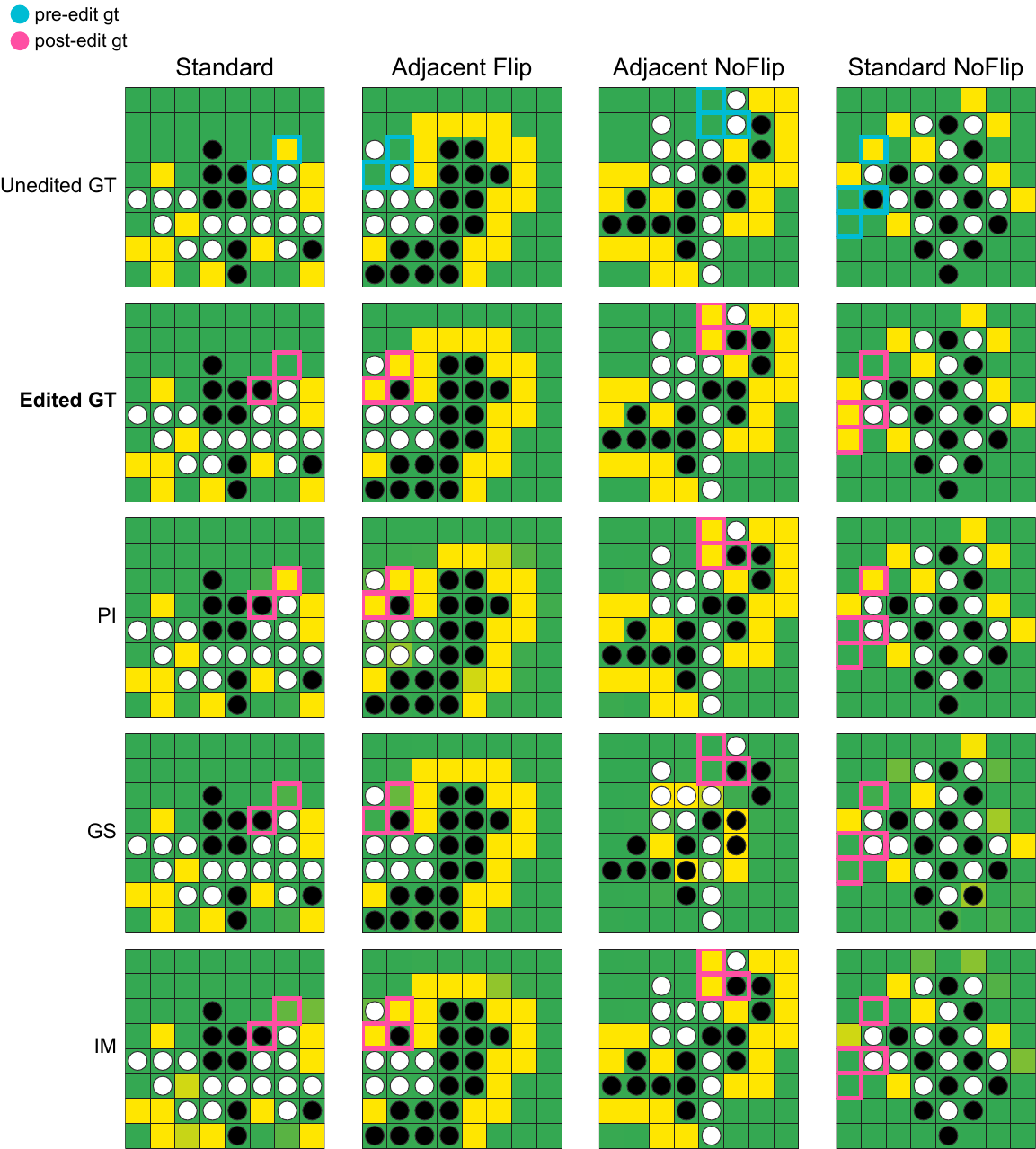}
\caption{Additional qualitative visualization of Othello edits.}
\label{fig:more_qualitative_edits_othello_1}
\end{figure}
\begin{figure}[ht]
\centering
\includegraphics[width=1.0\linewidth]{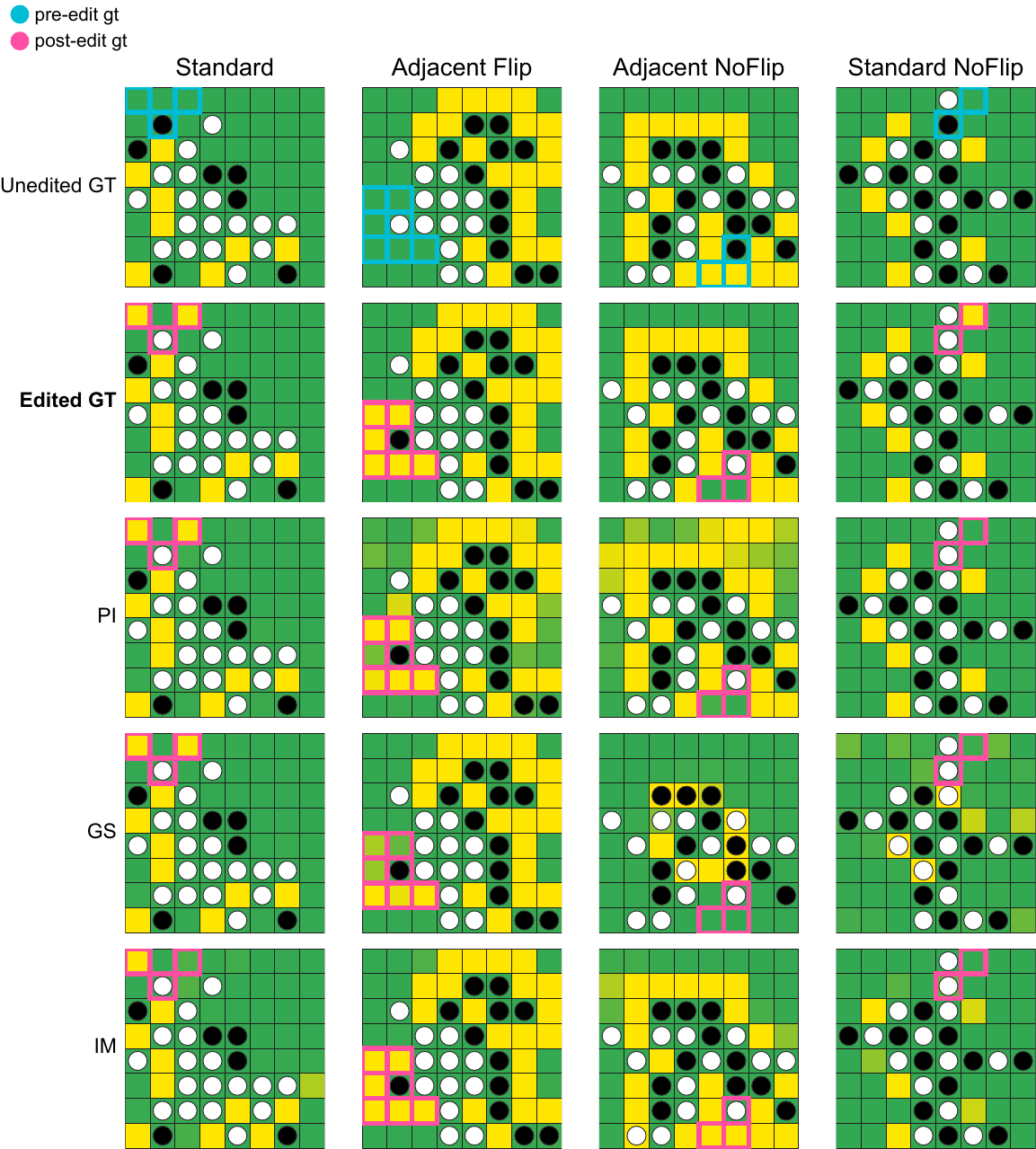}
\caption{Additional qualitative visualization of Othello edits.}
\label{fig:more_qualitative_edits_othello_2}
\end{figure}
\begin{figure}[ht]
\centering
\includegraphics[width=1.0\linewidth]{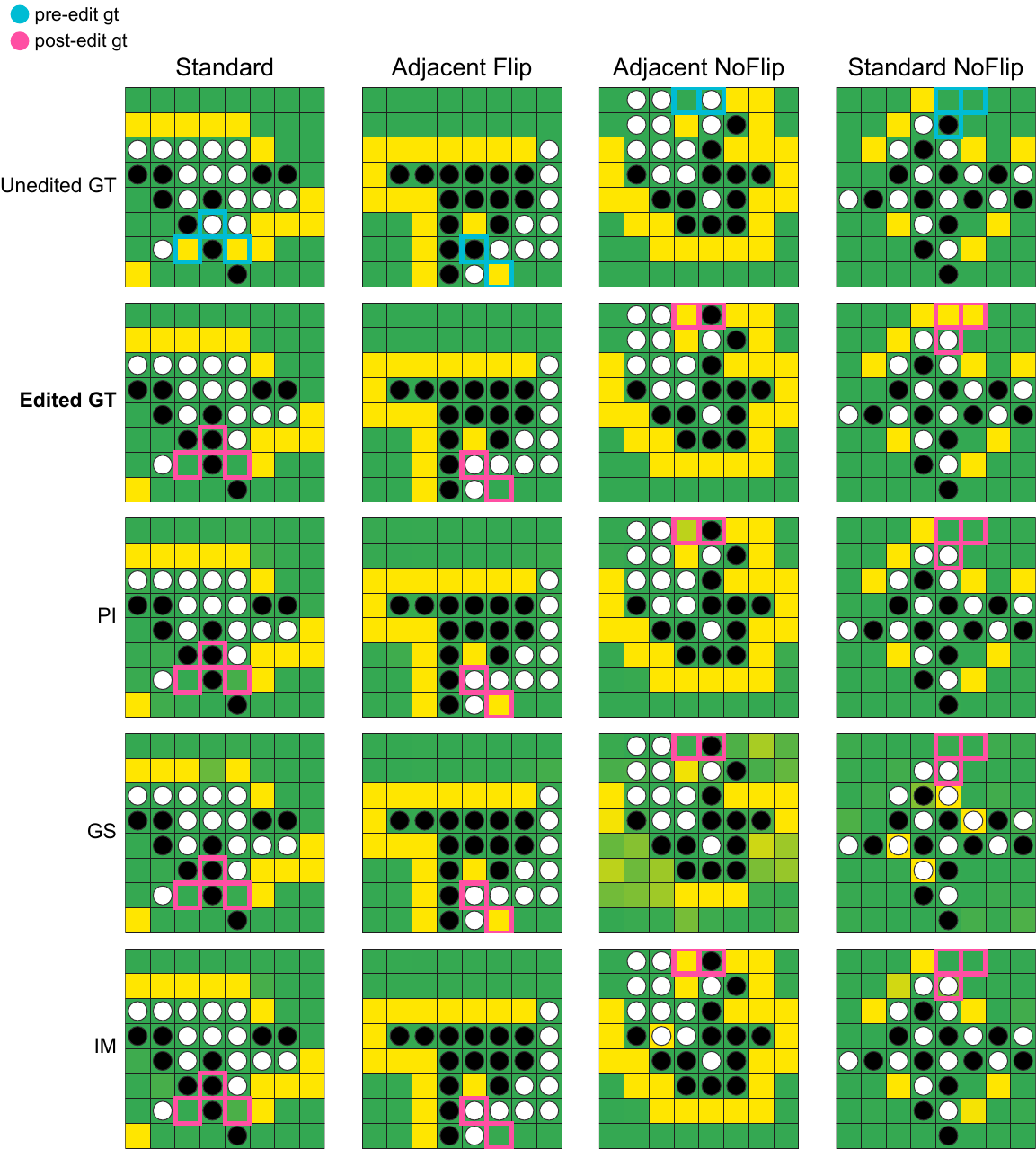}
\caption{Additional qualitative visualization of Othello edits.}
\label{fig:more_qualitative_edits_othello_3}
\end{figure}
\begin{figure}[ht]
\centering
\includegraphics[width=1.0\linewidth]{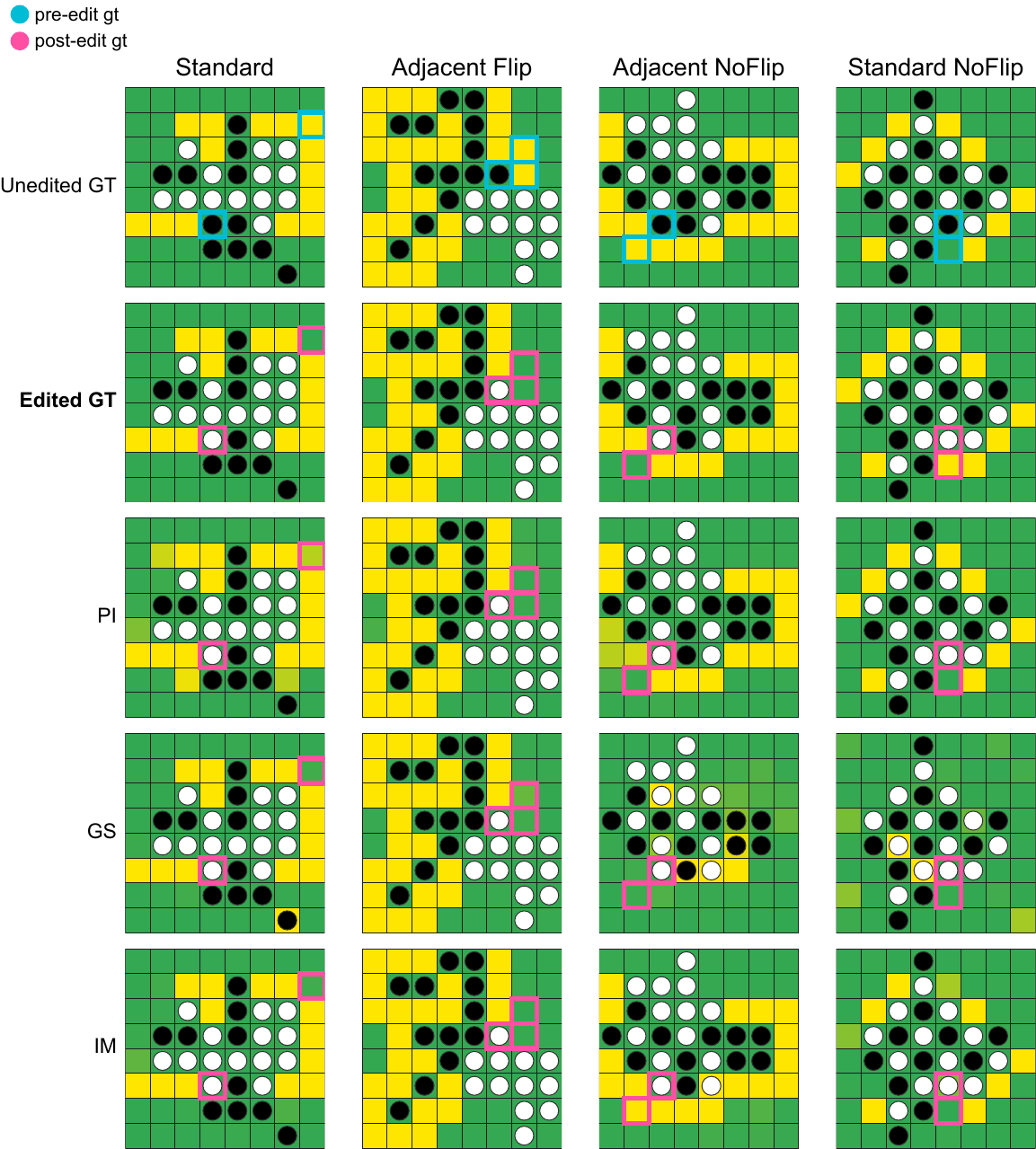}
\caption{Additional qualitative visualization of Othello edits.}
\label{fig:more_qualitative_edits_othello_4}
\end{figure}
\begin{figure}[ht]
\centering
\includegraphics[width=1.0\linewidth]{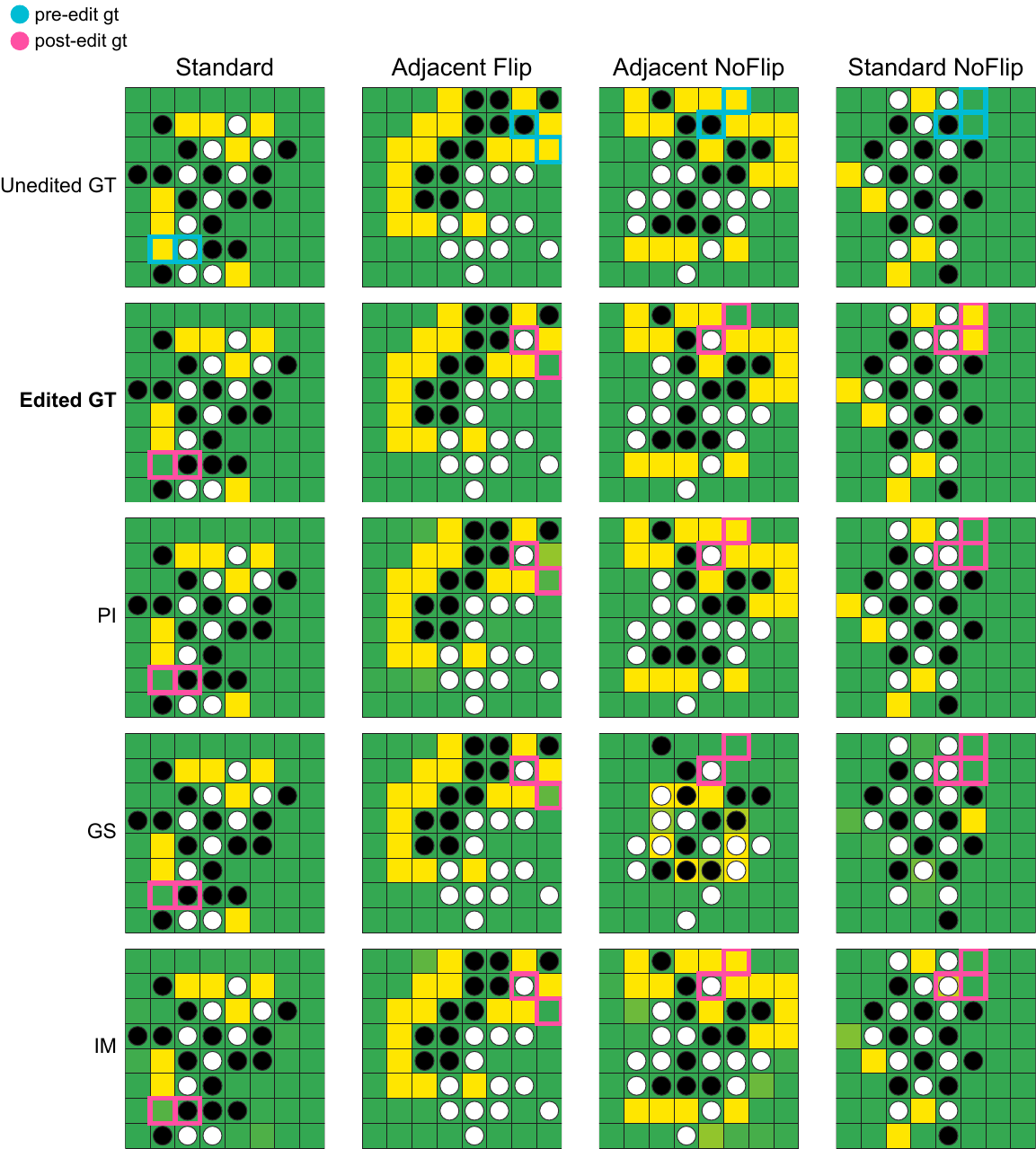}
\caption{Additional qualitative visualization of Othello edits.}
\label{fig:more_qualitative_edits_othello_5}
\end{figure}

\begin{figure}[ht]
\centering
\includegraphics[width=1.0\linewidth]{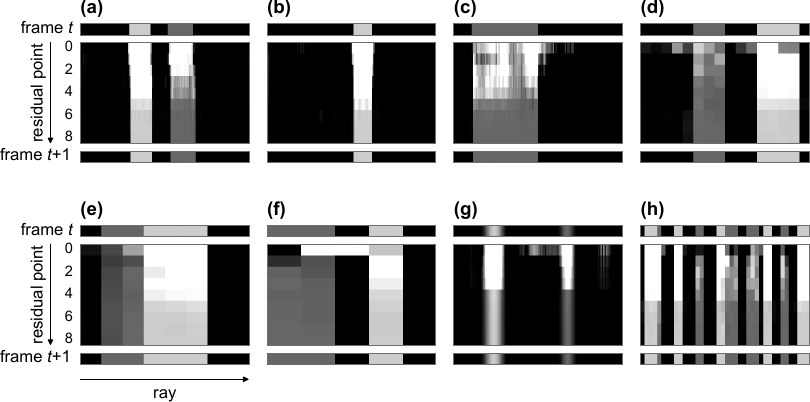}
\caption{\rev{The latent state at each residual point of the (a) \texttt{standard}, (b) \texttt{blink}, (c) \texttt{128-ray}, (d) \texttt{16-ray}, (e) \texttt{8-ray}, (f) \texttt{5-ray}, (g) \texttt{smooth}, and (h) \texttt{obs5} Rayworld models, passed directly through the decoders, for held-out sequences (\cref{app:inverse_controls}).}}
\label{fig:decoder_readout}
\end{figure}

\FloatBarrier
\begin{revblock}
\subsection{Controls for the Inverse Map}
\label{app:inverse_controls}
The inverse map could succeed without using the model's own representations, for instance by learning to edit the most recent input at the embedding layer. We test this on \texttt{standard} Rayworld (\cref{tab:inverse_controls}).

\paragraph{Splicing frames.}
We replace the last frames of the history with renders of the edited world, found by running the edited state backward as in \cref{app:history_editing}. One spliced frame fails, two edit moderately, and three match the inverse map. A learned renderer, an MLP of the inverse map's size trained on the same pairs, falls short when its frames are spliced in, reaching only $+0.46$ even with ten frames. So the inverse map is not just modifying the model's input context. Scored on its own outputs, the renderer reaches an Edit Index close to the inverse map's, but its loss is about ten times the Bayes floor, against about 1.3 times for the transformer (\cref{tab:predictive_skill}).

\paragraph{Patching in the counterfactual latent state.}
We run the model on the edited world's full history and write its latent state at one residual point, which is what the inverse map learns to predict. This fails at the embedding, where it amounts to replacing the current frame, succeeds by layer 1, and edits the prediction almost exactly from the middle of the network on. The inverse map follows a similar profile.

\paragraph{Reading each point through the decoder.}
\Cref{fig:decoder_readout} passes the latent state at each residual point directly through the decoder $G_D$. A rough version of the next frame appears by point 3, and most of the final rendering by point 5, which keeps improving afterward. The inverse map edits well by point 3 and plateaus around point 5, so it succeeds once a partial rendering has formed. 
\end{revblock}

\begin{table}[ht]
\centering
\small
\setlength{\tabcolsep}{4pt}

(a) Each control at its best setting\\[0.35em]
\begin{tabular}{l EF}
\hline
\noalign{\vspace{0.25em}}
Edit & \multicolumn{1}{c}{Index} & \multicolumn{1}{c}{Fid.} \\
\hline
\noalign{\vspace{0.1em}}
Splice the last frame (equal to patching from that run) & -0.09 & 0.37 \\
Splice the last two frames & +0.36 & 0.55 \\
Splice the last three frames & +0.63 & 0.68 \\
Learned renderer, two frames spliced & -0.22 & 0.28 \\
Learned renderer, three frames spliced & +0.09 & 0.41 \\
Learned renderer, ten frames spliced & +0.46 & 0.46 \\
Learned renderer, scored on its own outputs & +0.61 & 0.66 \\
Inverse map & +0.59 & 0.66 \\
Counterfactual patch & +0.91 & 0.84 \\
\hline
\end{tabular}

\vspace{1em}

(b) The counterfactual patch and the inverse map at every residual point\\[0.35em]
\begin{tabular}{l EF G EF}
\hline
\noalign{\vspace{0.25em}}
& \multicolumn{2}{c}{Counterfactual patch} & & \multicolumn{2}{c}{Inverse map} \\
Residual point & \multicolumn{1}{c}{Index} & \multicolumn{1}{c}{Fid.} & & \multicolumn{1}{c}{Index} & \multicolumn{1}{c}{Fid.} \\
\hline
\noalign{\vspace{0.1em}}
0 (embedding) & -0.09 & 0.37 & & -0.23 & 0.25 \\
1 & +0.41 & 0.65 & & +0.06 & 0.43 \\
2 & +0.59 & 0.75 & & +0.29 & 0.54 \\
3 & +0.77 & 0.82 & & +0.44 & 0.61 \\
4 & +0.86 & 0.84 & & +0.54 & 0.67 \\
5 & +0.89 & 0.84 & & +0.57 & 0.66 \\
6 & +0.90 & 0.84 & & +0.59 & 0.66 \\
7 & +0.90 & 0.84 & & +0.58 & 0.64 \\
8 & +0.91 & 0.84 & & +0.55 & 0.60 \\
\hline
\end{tabular}
\caption{\rev{Controls for the inverse map on \texttt{standard} Rayworld (\cref{app:inverse_controls}). Panel (a) gives each control at its best setting, and panel (b) scores the counterfactual patch and the inverse map through every residual point. Columns as in \cref{tab:editability}.}}
\label{tab:inverse_controls}
\end{table}

\FloatBarrier
\begin{revblock}
\subsection{Steering Along Encoding Directions and at Every Point}
\label{app:steering_variants}
PI moves the latent state along the pseudoinverse of the linear probe. A probe's weights need not point the way the latent state actually moves when the simulator state changes \citep{haufe2014interpretation}, so we also steer along that direction directly. At each residual point, we measure from the probe's data how much the average latent state shifts per unit change in each state variable, by regressing the latent state on the simulator state. For a single binary variable, this is the difference of class means used in contrastive steering \citep{rimsky2024caa}. This editor makes the same edit as PI with these directions in place of the pseudoinverse. We sweep both over the same residual points and step sizes, writing once or at every later point as GS does. In \texttt{standard} Rayworld, none of these succeeds: the best reaches an Edit Index of $-0.02$ at an Edit Fidelity of $0.02$, against $+0.59$ at $0.66$ for the inverse map, including allowing standard PI to edit across multiple layers.
\end{revblock}

\FloatBarrier
\begin{revblock}
\subsection{Robustness Checks and Fitting at the Edit Step}
\label{app:robustness}

\paragraph{Held-out setting selection.}
The main tables choose each editor's residual point and step size on the same 1000 cases they report. Choosing them on a separate validation set instead picks the same setting in 87\% of cells, and changes no Edit Index or Edit Fidelity by more than $0.1$.

\paragraph{One-ray edits.}
The Rayworld benchmarks keep only teleports that change at least two rays. Keeping the one-ray teleports changes no conclusion, and the trends with ray count remain.

\paragraph{Fitting at a single time step.}
The inverse map and the probes are fit on all positions of each sequence, and so is the $R^2$ in \cref{tab:decodability}. Refitting the inverse map on the edit step alone changes its Edit Index by at most $0.12$ on every variant except \texttt{adjacent-noflip}. In Rayworld its $R^2$ rises only modestly on the three 128-ray variants and stays below $0.4$, but rises on the coarser variants, to near $1$ on \texttt{5-ray}.

\paragraph{\texttt{adjacent-noflip}.}
Fit at the edit step, the inverse map succeeds at an Edit Index of $+0.86$ with an Edit Fidelity of $0.75$, even though no game could reach these boards (\cref{app:legal_vs_illegal_othello}). Fit across all time steps it fails, although it fits the latent state with an $R^2$ of up to $0.98$, and it also fails when the board is described by absolute color instead of mine and theirs (Edit Index $+0.06$). \Cref{tab:step_transfer} applies the map fit at step 20 to edits made at other steps. It works at nearby even steps, less well with distance, and fails at every odd step, where the other player moves. Passes, which would break this link between step and player, average 1.11 per game and almost never occur before step 20. Probes fit at the edit step show much less gain. GS still fails, and PI reaches $+0.26$ with an Edit Fidelity of $0.15$, below the index of $+0.35$ on \texttt{adjacent-flip} without the single-step fitting. So \texttt{adjacent-noflip} can be edited under the right circumstances, but not robustly, unlike \texttt{standard} and \texttt{adjacent-flip}, where a single inverse map fit across all time steps succeeds.
\end{revblock}

\begin{table}[ht]
\centering
\small
\setlength{\tabcolsep}{4pt}
\begin{tabular}{l EEEEEEEEE}
\hline
\noalign{\vspace{0.25em}}
Step of the edit & \multicolumn{1}{c}{16} & \multicolumn{1}{c}{17} & \multicolumn{1}{c}{18} & \multicolumn{1}{c}{19} & \multicolumn{1}{c}{20} & \multicolumn{1}{c}{21} & \multicolumn{1}{c}{22} & \multicolumn{1}{c}{23} & \multicolumn{1}{c}{24} \\
\hline
\noalign{\vspace{0.1em}}
Edit Index & +0.73 & -0.24 & +0.86 & -0.22 & +0.88 & -0.27 & +0.73 & -0.30 & +0.44 \\
Edit Fidelity & +0.53 & -3.89 & +0.75 & -4.07 & +0.77 & -4.29 & +0.67 & -4.28 & +0.39 \\
\hline
\end{tabular}
\caption{\rev{The inverse map fit on step 20 of \texttt{adjacent-noflip}, scored on single-flip edits made at nearby steps (\cref{app:robustness}). It works at even steps and fails at odd ones, where the other player moves.}}
\label{tab:step_transfer}
\end{table}

\FloatBarrier
\begin{revblock}
\subsection{Withholding the Viewpoint in Rayworld}
\label{app:pose_delay}
The principle of \cref{sec:forcing_state} can be pictured in the following way. Think of the causal state as a sheet wrapped loosely around the simulator's state. It follows the broad outline but not every detail. Anything that refines the causal states draws air out, like vacuum sealing, so the sheet conforms more closely to the simulator's state. But the sheet can never pass inside the simulator, since the simulator's state is always sufficient. Because the causal state is all a model must learn, a tighter sheet forces the model to represent more of the simulator's state. This section tests the idea in Rayworld, by making the model hold a state that must serve viewpoints it does not yet know, which provably refines the causal states.

\paragraph{Setup.}
Each Rayworld frame is rendered from a different viewpoint, and the model receives the viewpoint to render from only after taking one step without knowing it. The sequence alternates frame positions and viewpoint positions. A frame position reads a frame, and the viewpoint position after it reads the next viewpoint and predicts the frame rendered from it. In the one-step model, the viewpoint position attends only to the frame position before it. That frame position must therefore commit to a latent state before it knows the next viewpoint, and hold everything needed to render the next frame from any viewpoint. In the full-history model, the viewpoint position attends to every earlier position. Probes and editors act on the frame positions (\cref{sec:experimental_setup}).

\paragraph{Results.}
With one step of attention, the Edit Index of PI and GS rises on \texttt{8-ray}, with a smaller gain on \texttt{128-ray} (\cref{tab:pose_delay}). The inverse map edits only when it is written at several residual points at once, and it then edits very well, at an Edit Index of $+0.84$. With full-history attention, the gain disappears, as the theory predicts. The viewpoint position can then read the whole history once the viewpoint is known, so no state formed before the viewpoint has to summarize the history. For the state that does the predicting, the viewpoint is part of the history rather than a future input. The future inputs it must serve collapse to the single viewpoint already given, and the refinement is lost. Probe steering also fails on \texttt{obs5} (\cref{app:additional_rayworld_runs}) for the same reason. Its five viewpoints make the positions easier to infer, but they are fixed and shown in every frame, so the model never has to prepare for a viewpoint it does not yet know.
\end{revblock}

\begin{table}[ht]
\centering
\small
\setlength{\tabcolsep}{4pt}
\begin{tabular}{ll EF G EF}
\hline
\noalign{\vspace{0.25em}}
& & \multicolumn{2}{c}{PI} & & \multicolumn{2}{c}{GS} \\
Model & Viewpoint &
\multicolumn{1}{c}{Index} & \multicolumn{1}{c}{Fid.} & &
\multicolumn{1}{c}{Index} & \multicolumn{1}{c}{Fid.} \\
\hline
\noalign{\vspace{0.1em}}
\multirow{3}{*}{\texttt{8-ray}}
& single observer        & +0.21 & 0.00 & & -0.07 & 0.12 \\
& withheld, full history & -0.79 & 0.00 & & -0.61 & 0.01 \\
& withheld, one step     & +0.55 & 0.41 & & +0.42 & 0.37 \\
\cline{1-7}
\noalign{\vspace{0.1em}}
\multirow{2}{*}{\texttt{128-ray}}
& single observer        & -0.02 & 0.02 & & -0.10 & 0.03 \\
& withheld, one step     & +0.20 & +0.14 & & +0.14 & +0.15 \\
\hline
\end{tabular}
\caption{\rev{Editability of continuous positions when the model must commit to a state before it knows the next viewpoint (\cref{sec:forcing_state}). With one step of attention, the next step reads only that state, and with full history it reads every earlier position. Single-observer rows repeat \cref{tab:editability}. The inverse map is discussed in \cref{app:pose_delay}.}}
\label{tab:pose_delay}
\end{table}

\begin{revblock}
\paragraph{Refinement toward the simulator.}
Write $\sim_u$ for the relation of \cref{eq:causal_state} restricted to one future input sequence $u = u_{>t}$, and ${\sim_{\mathcal{U}}} = \bigcap_{u \in \mathcal{U}} {\sim_u}$ for a set $\mathcal{U}$ of allowed input sequences. If $\mathcal{U} \subseteq \mathcal{U}'$, every condition of $\sim_{\mathcal{U}}$ is also a condition of $\sim_{\mathcal{U}'}$, so the causal states under $\mathcal{U}'$ refine those under $\mathcal{U}$. With one fixed viewpoint, as in our other Rayworld variants, $\mathcal{U}$ holds a single input sequence, and two histories need only agree on what that one viewpoint will show. A viewpoint the model does not yet know makes it consider all possible views, which separates histories that one view would equate.

The simulator bounds this refinement. Its state $s_t$, together with the future inputs, fixes the distribution of future observations, and the inputs we draw are independent of it. Two histories with the same posterior $P(s_t \mid x_{\le t})$ over the simulator's state therefore share a causal state for any $\mathcal{U}$. Causal states are thus coarse-grainings of this posterior, and allowing more inputs moves them toward it. In the vacuum picture, this posterior is the object the sheet is wrapped around. It is the most the history can say about the simulator's state, and when the observations pin that state down, the causal states approach the simulator's state itself.

The same reasoning applies to other kinds of input. An agent's future actions are inputs too, so a world model trained to predict under many possible actions faces the same pressure. Transformers trained on data from hidden Markov models represent such posteriors linearly in their residual stream \citep{shai2024transformers}, and generative query networks \citep{eslami2018neural} render a stored scene from viewpoints given afterward, a design this view suggests should better yield a state a simulator can set. Training a model to serve inputs it does not know in advance may be a general way to make it hold such a state.
\end{revblock}

\end{document}